\documentclass[runningheads]{llncs}
\usepackage{float}

\usepackage{eccv}

\usepackage{eccvabbrv}

\usepackage{graphicx}
\usepackage{booktabs}

\usepackage[accsupp]{axessibility}  

\usepackage[dvipsnames]{xcolor}
\usepackage{times}
\usepackage{epsfig}
\usepackage{graphicx}
\usepackage{amsmath}
\usepackage{amssymb}
\usepackage{multirow}
\usepackage{amssymb}
\usepackage{pifont}

\usepackage{adjustbox}

\usepackage{colortbl}
\usepackage{xcolor}
\usepackage{graphicx}
\definecolor{globalcolor}{RGB}{255, 102, 0}
\definecolor{localcolor}{RGB}{0, 153, 76} 
\definecolor{lightgray}{gray}{0.9}

\usepackage{tcolorbox}
\usepackage{listings}
\usepackage{pict2e} 

\tcbuselibrary{breakable}

\usepackage{subcaption} 

\DeclareMathSymbol{@}{\mathord}{letters}{"3B}
\usepackage{adjustbox}
\usepackage{algorithm}
\usepackage{algpseudocode}
\usepackage{arydshln}
\usepackage{times}
\usepackage{placeins}
\usepackage{float}

\usepackage{wrapfig}

\usepackage{hyperref}

\usepackage{orcidlink}

\begin{document}

\title{\textcolor{blue}{KHMP}: Frequency-Domain \textcolor{blue}{K}alman Refinement for High-Fidelity \textcolor{blue}{H}uman \textcolor{blue}{M}otion \textcolor{blue}{P}rediction}

\titlerunning{KHMP}

\author{Wenhan Wu\inst{1,2}\and
Zhishuai Guo\inst{3} \and
Chen Chen\inst{4} \and
Srijan Das\inst{2} \and
Hongfei Xue\inst{2} \and
Pu Wang\inst{2} \and
Aidong Lu\inst{2}}

\authorrunning{W.~Wu et al.}

\institute{School of Engineering, Yunnan University, Kunming, Yunnan, China \and 
Department of Computer Science, University of North Carolina at Charlotte, Charlotte, NC, USA\\
\email{\{wwu25, sdas24, hongfei.xue, pu.wang, aidong.lu\}@charlotte.edu} \and 
Department of Computer Science, Northern Illinois University, DeKalb, IL, USA\\
\email{zguo@niu.edu} \and
Institute of Artificial Intelligence, University of Central Florida, Orlando, FL, USA\\
\email{chen.chen@ucf.edu}}
\maketitle

\begin{abstract}
Stochastic human motion prediction aims to generate diverse, plausible futures from observed sequences. Despite advances in generative modeling, existing methods often produce predictions corrupted by high-frequency 
jitter and temporal discontinuities. To address these challenges, we introduce \textbf{KHMP}, a novel framework featuring an adaptive \textbf{K}alman filter applied in the DCT domain to generate high-fidelity 
\textbf{h}uman \textbf{m}otion \textbf{p}redictions. By treating high-frequency DCT coefficients as a frequency-indexed noisy signal, the Kalman filter recursively suppresses noise while preserving motion 
details. Notably, its noise parameters are dynamically adjusted based on estimated Signal-to-Noise Ratio (SNR), enabling aggressive denoising for jittery predictions and conservative filtering for clean motions. 
This refinement is complemented by training-time physical constraints (temporal smoothness and joint angle limits) that encode biomechanical principles into the generative model. Together, these innovations establish a new paradigm integrating adaptive signal processing with physics-informed learning. Experiments on the Human3.6M and HumanEva-I datasets demonstrate that KHMP achieves state-of-the-art accuracy, effectively mitigating jitter artifacts to produce smooth and physically plausible motions. Our project is publicly available at: \textcolor{magenta}{\url{https://github.com/wenhanwu95/KHMP-Project-Page.git}}.
\end{abstract}

\section{Introduction}
\label{sec:intro}



Stochastic human motion prediction (SHMP) aims to forecast a diverse set of plausible future movements from an observed sequence. To model the complex, high-dimensional distribution of human dynamics, generative approaches~\cite{DLow, STARS, SLD-HMP, Belfusion, MotionDiff, SkeletonDiff, CoMotion} have emerged as the predominant paradigm, leveraging their tractable latent spaces to capture movement variability. However, despite their success in generating diverse samples, designing a standalone generative framework that concurrently ensures high-fidelity, temporally coherent, and physically plausible trajectories remains a critical challenge.

Specifically, existing stochastic prediction models face two fundamental limitations. First, they \textbf{lack a frequency-aware, adaptive mechanism that can selectively suppress high-frequency jitter while preserving low-frequency motion dynamics} (\textcolor{blue}{\textbf{Challenge 1}}). The stochastic sampling inherent to latent generative spaces frequently introduces high-frequency unstructured noise into the predicted trajectories. Although some recent approaches~\cite{MotionDiff, MotionWavelet} explore frequency-domain representations, they typically rely on general-purpose architectures that uniformly process all frequency components. This one-size-fits-all approach cannot adapt to the varying levels of sampling noise across different stochastic predictions, often failing to suppress jitter effectively without compromising natural motion details.

\begin{wrapfigure}{r}{0.5\textwidth}
  \vspace{-25pt}
  \centering
  \includegraphics[width=\linewidth]{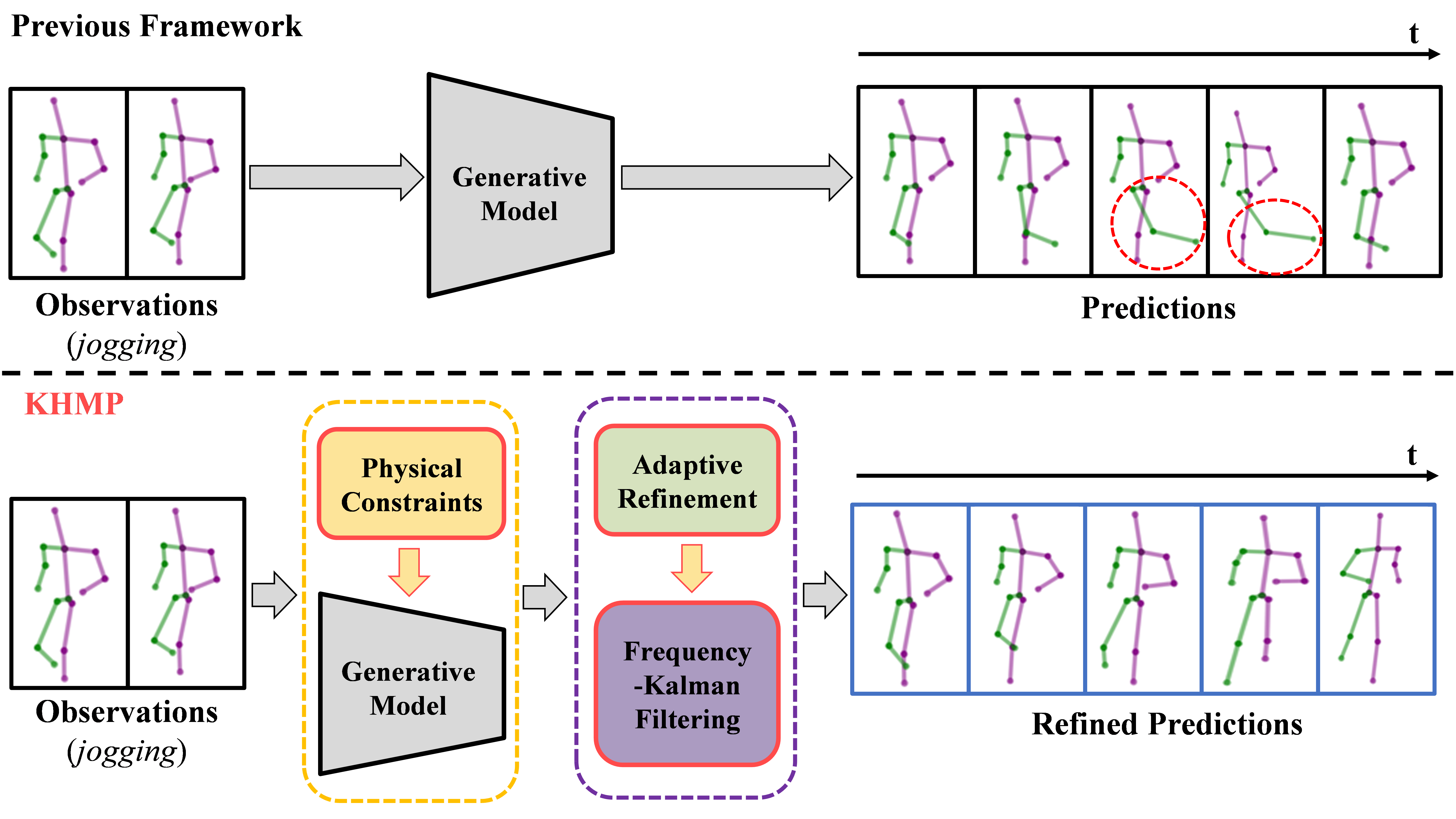}
  \caption{Comparison illustrating motion prediction for the \textit{jogging} action. Previous frameworks (top) often generate predictions exhibiting physical implausibility (highlighted by red circles) and temporal discontinuity (abrupt pose transitions). \textbf{KHMP} (bottom) integrates physical constraints during training and adaptive frequency-Kalman refinement during inference to produce more plausible and temporally smoother motion sequences.}
  \label{fig:fig1}
  \vspace{-15pt}
\end{wrapfigure}

Second, current SHMP models \textbf{primarily rely on frame-wise reconstruction losses without explicitly enforcing biomechanical laws governing human motion} (\textcolor{blue}{\textbf{Challenge 2}}). Although minimizing per-frame errors ensures statistical similarity, the absence of structural priors can lead to subtle yet perceptually jarring artifacts, such as unnatural jitter, anatomically impossible poses, and physically inconsistent bone deformations~\cite{SLD-HMP, SkeletonDiff}.

In this work, we introduce \textbf{KHMP}, a novel framework that directly addresses these limitations. \textcolor{blue}{\textbf{To tackle Challenge 1}}, we introduce a refinement mechanism leveraging the insight that high-frequency jitter resembles structured noise in the motion's spectral representation. Traditional methods either ignore this or apply hard frequency masking, which indiscriminately destroys fine-grained motion details. Instead, recognizing the Kalman filter~\cite{kalman1960new} as the optimal linear estimator for recursive denoising, we propose applying it directly within the Discrete Cosine Transform (DCT~\cite{ahmed1974discrete}) frequency domain. Our core innovation lies in treating the high-frequency DCT coefficients of the predicted 3D joint sequences as a frequency-indexed sequence. By modeling a first-order Gaussian-Markov relationship between adjacent frequency components, the filter recursively smooths the high-frequency spectrum, which typically represents jittering noise. Furthermore, because noise levels vary substantially across stochastic predictions, a fixed filter is suboptimal. We thus introduce an adaptive strategy where the Kalman noise parameters (process and observation covariance) are dynamically adjusted based on an estimated Signal-to-Noise Ratio (SNR). Derived from the high-frequency energy ratio, this mechanism enables aggressive denoising for highly jittery outputs while conservatively preserving the delicate dynamics of clean predictions.

\textcolor{blue}{\textbf{To address Challenge 2}}, we augment the generative training objective with structured physical constraints to complement our Frequency-Kalman refinement. By incorporating a temporal smoothness loss penalizing frame-to-frame displacement variance, alongside a soft cosine-based joint angle constraint, we explicitly guide the learned latent distribution toward anatomically plausible and biomechanically valid motions. An overview of our \textit{novel} KHMP is illustrated in Fig.~\ref{fig:fig1}, and the key contributions are:

\begin{enumerate}
    \item \textit{We propose the \textbf{first} application of the Kalman filter in the frequency domain (DCT)} to treat high-frequency coefficients as an indexed sequence in SHMP, recursively suppressing jitter artifacts while preserving essential motion dynamics.
    \item \textit{An adaptive SNR-based parameterization strategy} that dynamically adjusts the Kalman filter's process and observation noise covariances, ensuring optimal smoothing levels across diverse prediction qualities.
    \item \textit{A unified training framework augmented with physical constraints} (temporal smoothness and joint angle limits) that explicitly encodes biomechanical priors into the generative process to ensure anatomical validity.
    \item Extensive experiments on standard benchmarks (Human3.6M and HumanEva-I) demonstrating that KHMP achieves state-of-the-art accuracy, effectively resolving the accuracy-diversity trade-off by producing highly faithful and temporally coherent motions.
\end{enumerate}

\section{Related Work}
\label{sec:related}
\vspace{-0.05in} 
\noindent\textbf{Stochastic Human Motion Prediction.}
To capture the inherent multimodality of human motion prediction, the field has progressed from early deterministic models \cite{ERD, acLSTM} towards stochastic generative frameworks. Initially dominated by Variational Autoencoders (VAEs) and Generative Adversarial Networks (GANs), a primary research thrust involved structuring the latent space to foster diversity. This was pursued through various techniques, such as normalizing flows \cite{DLow}, specialized sampling objectives \cite{BoM, DivSamp}, and the design of explicit latent bases, including orthogonal semantic directions \cite{SLD-HMP} and spatial-temporal anchors \cite{STARS}. More recently, diffusion models have become the predominant paradigm, achieving state-of-the-art results by framing prediction as a conditional denoising process. This approach has been successfully instantiated through diverse architectural designs, from behavior-driven latent diffusion \cite{Belfusion} to Transformer-based models \cite{TransFusion, SwiftDiff} and explorations into non-isotropic noise schedules \cite{SkeletonDiff}.

However, despite the generative power of both VAE and diffusion-based approaches, a common thread is their primary reliance on direct frame-wise supervision (e.g., an L2 loss in the pose space). This often prioritizes statistical similarity over signal-level fidelity, leading to outputs that can still suffer from subtle, yet perceptually significant artifacts, such as high-frequency jitter and temporal discontinuities. Our work addresses this persistent gap by proposing a novel Frequency-Kalman filter for temporal smoothing with structured physical loss functions, providing a complementary module that enhances the predictions.

\noindent\textbf{Kalman Methods.}
The Kalman filter is a seminal algorithm for state estimation in linear dynamical systems~\cite{kalman1960new}. In human motion analysis, its traditional application has been in the \textit{time domain}, primarily to track or smooth noisy motion capture data ~\cite{sul1998synthesis, liu2010human, zhou2022joint, LannanTKFAE2022, LugrisRealtime2023, liu2024recurrent}. For instance, Lugrís et al.~\cite{LugrisRealtime2023} used a time-domain EKF for real-time kinematic reconstruction, including accelerations aimed at musculoskeletal analysis, while Lannan et al.~\cite{LannanTKFAE2022} combined a Tobit Kalman Filter with an autoencoder via latent space optimization for motion enhancement. These classical approaches model the temporal evolution of joint positions directly, often assuming simplified dynamics or focusing on latent space correction rather than frequency-specific refinement. More advanced variants include~\cite{liu2024recurrent}, which employs an unscented Kalman filter (UKF) to forecast muscle-force inputs for a dynamic model that predicts human arm motion. 
In parallel, recent work on general inverse problems has revitalized Kalman-based methods in derivative-free settings, such as Ensemble Kalman Inversion (EKI)~\cite{iglesias2013ensemble, zheng2025ensemble}.

Our work departs from both traditional time-domain filtering and modern ensemble-based guidance. We propose a novel paradigm by applying a recursive Kalman filter in the \textit{frequency domain}. Instead of modeling temporal dynamics, we model the relationships between adjacent DCT coefficients. The key innovation lies in making this filter adaptive to the signal's spectral properties, a targeted refinement strategy distinct from prior methods.
\section{Preliminaries}
\label{sec:preliminaries}

\subsection{Problem Formulation}
\label{sec:problem formulation}
Given an observed human motion sequence $\mathbf{X}_{1:T'} = \{\mathbf{x}_1, \dots, \mathbf{x}_{T'}\}$, where each $\mathbf{x}_t \in \mathbb{R}^{J \times 3}$ denotes the 3D coordinates of $J$ joints at time $t$, the goal is to generate plausible future motion trajectories $\widehat{\mathbf{Y}}_{1:T'}$ for a prediction horizon of $T'$ frames. Our proposed refinement operates as a post-processing module on generated motion sequences. We integrate it with a VAE-based architecture~\cite{SLD-HMP} and demonstrate substantial improvements in prediction quality.

\subsection{Kalman Methodology}
The Kalman filter \cite{kalman1960new, krishnan2015deep} is a recursive estimator for linear dynamical systems with Gaussian noise. It models the hidden-state evolution and noisy observations as:
\begin{align}
x_t &= A x_{t-1} + w_t, \quad w_t \sim \mathcal{N}(0, Q) \\
z_t &= H x_t + v_t, \quad v_t \sim \mathcal{N}(0, R)
\end{align}
where $x_t$ is the latent state at time $t$, $z_t$ is the observed signal, $A$ is the state transition matrix, $H$ is the observation matrix, and $Q$, $R$ are the process and the observation noise covariances, respectively.

The Kalman filter estimates the state posterior mean $\hat{x}_{t|t}$ and covariance $P_{t|t}$ through two steps:

\textbf{Prediction Step:}
\begin{align}
\hat{x}_{t|t-1} &= A \hat{x}_{t-1|t-1} \\
P_{t|t-1} &= A P_{t-1|t-1} A^\top + Q
\end{align}

\textbf{Update Step:}
\begin{align}
K_t &= P_{t|t-1} H^\top (H P_{t|t-1} H^\top + R)^{-1} \\
\hat{x}_{t|t} &= \hat{x}_{t|t-1} + K_t (z_t - H \hat{x}_{t|t-1}) \\
P_{t|t} &= (I - K_t H) P_{t|t-1}
\end{align}

In our setting, \textit{we adapt the Kalman filter to operate in the frequency domain, motivated by the ability to suppress the high-frequency components that are primarily responsible for motion jitter}. Specifically, we apply it to sequences of DCT coefficients indexed by frequency $k$, treating adjacent frequency components as a first-order Gaussian-Markov process ($A = 1$, $H = 1$). Note that, unlike traditional Kalman filtering over time $t$, our recursion operates on frequency indices $k$. This yields a simplified update form:
\begin{align}
\hat{x}_{k|k} &= \hat{x}_{k|k-1} + K_k (z_k - \hat{x}_{k|k-1}), \\
K_k &= \frac{P_{k|k-1}}{P_{k|k-1} + R},
\end{align}
which ensures optimal denoising under the minimum mean square error (MMSE) criterion.
This foundation enables our frequency-domain refinement strategy, where Kalman filtering is applied adaptively based on the energy of each DCT frequency component.

\begin{figure*}[t]
  \centering
  \includegraphics[width=1.0\linewidth]{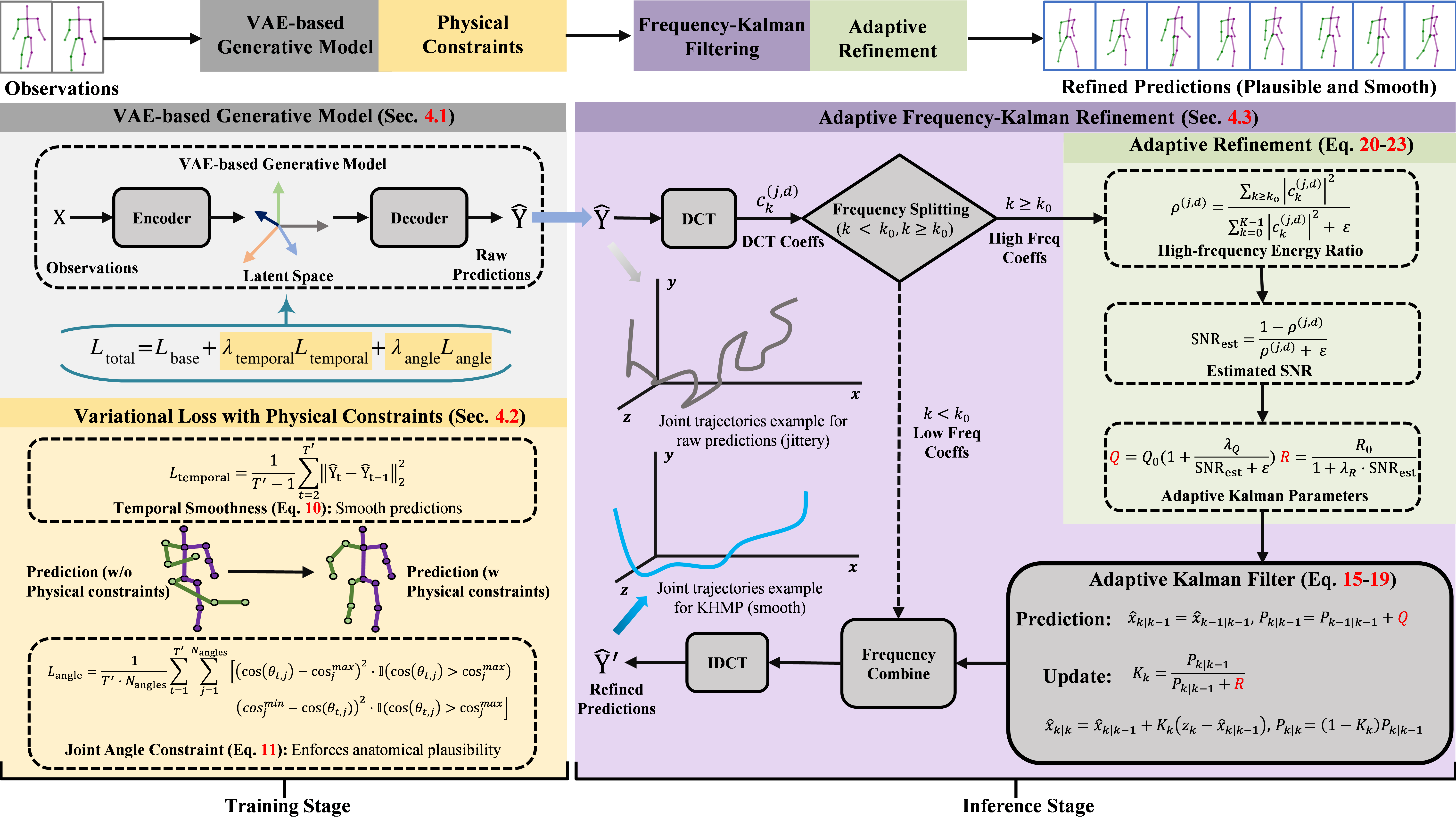}
    \caption{Overview of the proposed \textbf{KHMP} framework. During training (left), a VAE backbone model learns with standard losses augmented by structured \textbf{Physical Constraints} (Sec.~\ref{sec:loss}) for enhanced realism. During inference (right), raw predictions are processed by the \textbf{Adaptive Frequency-Kalman Refinement} module (Sec.~\ref{sec:freq_kalman}). This module uses \textbf{SNR}-based parameter adjustments and recursive Kalman filtering to adaptively smooth high-frequency noise, yielding refined predictions.}
    \label{fig:fig2}
\end{figure*}
\section{KHMP}
\label{sec:methods}
This section details KHMP, a novel method for generating physically plausible human motion sequences that are refined using a Kalman filter in the frequency domain.

\vspace{-0.02in} 
\subsection{Method Overview}
Our goal is to generate physically plausible and temporally coherent future motion sequences given an observed 3D pose sequence $\mathbf{X}_{1:T'} = \{\mathbf{x}_1, \dots, \mathbf{x}_{T'}\}$, where each frame $\mathbf{x}_t \in \mathbb{R}^{J \times 3}$ represents joint positions. The VAE backbone~\cite{SLD-HMP} serves as the generative engine (Sec. \ref{sec:problem formulation}), learning the underlying motion distribution and producing initial diverse predictions based on the input sequence. Enhanced by our integrated physical constraints during training (Sec. \ref{sec:loss}) and subsequent Frequency-Kalman refinement at inference (Sec. \ref{sec:freq_kalman}), our framework yields smoothed motion predictions $\widehat{\mathbf{Y}}' = \{\hat{y}'_1, \dots, \hat{y}'_{T'}\}$ that better adhere to realistic human motion dynamics.

To achieve this, we introduce a frequency-aware variational motion generation framework with physical constraints that incorporates innovative signal-level regularization and domain-level constraints. Our primary contribution lies in a novel DCT-based Kalman post-processing module that refines predicted high-frequency components. By modeling adjacent high-frequency DCT coefficients as a first-order Gaussian-Markov process along the frequency index, this module attenuates artifacts and improves temporal consistency, with an adaptive parameterization strategy that dynamically adjusts noise variances based on spectral energy. Additionally, we propose a unified training objective that incorporates novel physical consistency terms. Specifically, we introduce a temporal smoothness loss that penalizes frame-to-frame displacements to promote temporal coherence, and a joint-angle constraint loss that enforces anatomical plausibility via soft, cosine-based penalties against biomechanical limits.

The overall framework employs a variational motion generator refined by our frequency-aware module at inference, enabling high-fidelity motion prediction. The overall design is illustrated in Fig. \ref{fig:fig2}.

\subsection{Variational Loss with Physical Constraints}
\label{sec:loss}
To encourage temporally smooth and physically plausible motion generation, we augment the variational objective with two novel physical constraints: temporal smoothness and anatomical validity through joint angle limits.

\textbf{Temporal Smoothness.} To suppress abrupt motion changes, we apply a frame-wise temporal regularization term that penalizes large frame-to-frame displacements:
\begin{equation}
\mathcal{L}_{\text{temporal}} = \frac{1}{T'-1} \sum_{t=2}^{T'} \|\widehat{\mathbf{Y}}_t - \widehat{\mathbf{Y}}_{t-1}\|_2^2,
\end{equation}
where $\widehat{\mathbf{Y}}_t \in \mathbb{R}^{J \times 3}$ is the predicted pose at time $t$. This loss operates directly in the pose space during training, providing effective regularization for temporal coherence and continuous motion trajectories.

\textbf{Joint Angle Constraint.} To enforce anatomical plausibility, we constrain joint angles $\theta_{t,j}$ between parent and child bones at frame $t$ for each joint $j$. Existing generative frameworks, such as ~\cite{GSPS} and ~\cite{STARS}, directly calculate and penalize the raw angle values. However, deriving the exact angle requires the inverse cosine function ($\arccos$), whose gradient magnitude diverges as its input approaches $\pm 1$ (i.e., $\left|\frac{d}{dx}\arccos(x)\right| = \frac{1}{\sqrt{1-x^2}} \to +\infty$). Consequently, directly penalizing raw angle values introduces severe numerical instability for configurations that naturally arise in human motion (e.g., near-collinear body segments like fully extended limbs). To sidestep this risk, we instead directly constrain $\cos(\theta_{t,j})$, which is efficiently and safely computable via normalized dot products without any trigonometric inversions. We define physiologically valid ranges $[\cos^{\min}_j, \cos^{\max}_j]$ for the cosine value of joint angle. The angle loss is then defined as the average penalty over all timesteps and constraints:
\begin{equation}
\label{eq:angle_loss}
\resizebox{0.88\columnwidth}{!}{$\displaystyle
\begin{split}
\mathcal{L}_{\text{angle}} = \frac{1}{T' \cdot N_{\text{angles}}} \sum_{t=1}^{T'} \sum_{j=1}^{N_{\text{angles}}} \Big[
&(\cos(\theta_{t,j}) - \cos^{\max}_j)^2 \cdot \mathbb{I}(\cos(\theta_{t,j}) > \cos^{\max}_j) \\
&+ (\cos^{\min}_j - \cos(\theta_{t,j}))^2 \cdot \mathbb{I}(\cos(\theta_{t,j}) < \cos^{\min}_j)
\Big],
\end{split}
$}
\end{equation}
where $T'$ is the prediction horizon, $N_{\text{angles}}$ is the number of defined angle constraints, and $\mathbb{I}(\cdot)$ is the indicator function that equals 1 when the condition holds and 0 otherwise. This loss implements a soft penalty that remains zero for anatomically valid poses within $[\cos^{\min}_j, \cos^{\max}_j]$ and applies quadratic penalties when constraints are violated. By entirely avoiding the unstable $\arccos$ operation, our formulation prevents angles from becoming too small ($\cos(\theta_{t,j}) > \cos^{\max}_j$) or too large ($\cos(\theta_{t,j}) < \cos^{\min}_j$), strictly enforcing biomechanical plausibility while ensuring robust explosion-free differentiability during training. The detailed angle constraint settings are provided in \textcolor{magenta}{Supplementary Material (SupMat) H}.

\textbf{Unified Training Objective.} We train the model using a composite objective that combines the $\mathcal{L}_{\text{base}}$ from the baseline \cite{SLD-HMP} with our proposed physical constraints. The complete loss is:
\begin{equation}
\begin{aligned}
\mathcal{L}_{\text{total}} = &\ \mathcal{L}_{\text{base}} + \lambda_{\text{temporal}} \mathcal{L}_{\text{temporal}} + \lambda_{\text{angle}} \mathcal{L}_{\text{angle}},
\end{aligned}
\end{equation}
where $\mathcal{L}_{\text{base}}$ includes reconstruction loss ($\mathcal{L}_{\text{recon}}$), multimodal reconstruction ($\mathcal{L}_{\text{mm}}$), historical reconstruction ($\mathcal{L}_{\text{h}}$), diversity-promoting loss ($\mathcal{L}_{d}$), limb length consistency ($\mathcal{L}_{\text{limb}}$), and pose prior loss ($\mathcal{L}_{\text{nf}}$). For completeness, we provide their detailed definitions in \textcolor{magenta}{SupMat E}. Our \textit{key contribution} lies in introducing $\mathcal{L}_{\text{temporal}}$ and $\mathcal{L}_{\text{angle}}$, which encode temporal continuity and anatomical validity.

\subsection{Frequency-Kalman Refinement}
\label{sec:freq_kalman}

While the physical constraints introduced in Section~\ref{sec:loss} guide the model toward biomechanically valid predictions during training, they cannot effectively eliminate high-frequency jitter in individual samples, as the VAE's stochastic sampling inherently introduces variability in the latent space. To address this, \textit{we introduce a frequency-domain Kalman refinement module with an adaptive filtering strategy to mitigate residual high-frequency jitter and enhance temporal smoothness at inference}.

Given a predicted motion sequence $\widehat{\mathbf{Y}} \in \mathbb{R}^{T' \times J \times 3}$, we apply the Discrete Cosine Transform (DCT) along the temporal axis for each joint and coordinate, yielding frequency-domain coefficients $c_k^{(j,d)}$, where $k$ denotes the frequency index, $j$ the joint index, and $d \in \{x, y, z\}$ the coordinate channel.
We employ a two-stage frequency decomposition strategy, separating the spectrum into low-frequency ($k < k_0$) and high-frequency ($k \geq k_0$) components. For each high-frequency coefficient sequence, we apply recursive Kalman filtering along the frequency indices, treating them as a noisy signal to be refined:
\begin{align}
x_k &= x_{k-1} + w_k, \quad w_k \sim \mathcal{N}(0, Q), \label{eq:kalman_process} \\
z_k &= x_k + v_k, \quad v_k \sim \mathcal{N}(0, R), \label{eq:kalman_observation}
\end{align}
where $z_k$ denotes the $k$-th high-frequency DCT coefficient from the generator, modeled as a noisy observation of the underlying clean coefficient $x_k$, and $Q$, $R$ are the process and observation noise variances.

The Kalman filter recursively estimates $x_k$ using the standard prediction and update steps:
\begin{align}
\text{Prediction:} \quad & \hat{x}_{k|k-1} = \hat{x}_{k-1|k-1}, \label{eq:kalman_pred_state} \\
& P_{k|k-1} = P_{k-1|k-1} + Q, \label{eq:kalman_pred_cov} \\
\text{Update:} \quad & K_k = \frac{P_{k|k-1}}{P_{k|k-1} + R}, \label{eq:kalman_gain} \\
& \hat{x}_{k|k} = \hat{x}_{k|k-1} + K_k (z_k - \hat{x}_{k|k-1}), \label{eq:kalman_update_state} \\
& P_{k|k} = (1 - K_k) P_{k|k-1}. \label{eq:kalman_update_cov}
\end{align}

To enable a more principled, data-driven smoothing, we introduce an \textit{adaptive strategy} for the Kalman parameters based on an estimated Signal-to-Noise Ratio (SNR). First, for each channel $(j,d)$, we compute the high-frequency energy ratio $\rho^{(j,d)}$:
\begin{equation}
\label{eq:rho}
\rho^{(j,d)} = \frac{\sum_{k \geq k_0} |c_k^{(j,d)}|^2}{\sum_{k=0}^{K-1} |c_k^{(j,d)}|^2 + \epsilon},
\end{equation}
where $\epsilon = 10^{-8}$ for numerical stability. This ratio serves as a proxy for the noise level, from which we estimate the SNR for that channel as:
\begin{equation}
\label{eq:snr}
\text{SNR}_{\text{est}} = \frac{1 - \rho^{(j,d)}}{\rho^{(j,d)} + \epsilon}.
\end{equation}

The adaptive Kalman parameters are then set as non-linear functions of this estimated SNR:
\begin{align}
Q &= Q_0 \left(1 + \frac{\lambda_Q}{\text{SNR}_{\text{est}} + \epsilon}\right), \label{eq:adaptive_q} \\
R &= \frac{R_0}{1 + \lambda_R \cdot \text{SNR}_{\text{est}}}, \label{eq:adaptive_r}
\end{align}
where $Q_0, R_0$ are the base noise variances (with $R_0 \gg Q_0$, e.g., $R_0=10^{-2}, Q_0=10^{-6}$), and $\lambda_Q, \lambda_R$ are sensitivity hyperparameters. This formulation enables adaptive smoothing with a clear physical interpretation: the filter intelligently applies stronger smoothing to signals estimated as noisy (low SNR) while faithfully tracking signals estimated as clean (high SNR). As the signal quality improves (higher $\text{SNR}_{\text{est}}$), the observation noise $R$ decreases significantly (Eq.~\ref{eq:adaptive_r}), reflecting greater trust in the input coefficients $z_k$. Simultaneously, the process noise $Q$ also decreases (Eq.~\ref{eq:adaptive_q}), approaching $Q_0$ and enforcing a stronger smoothness prior. Crucially, due to $R_0 \gg Q_0$, $R$ remains the dominant factor in determining the Kalman gain $K_k = \frac{P_{k|k-1}}{P_{k|k-1} + R}$ (Eq.~\ref{eq:kalman_gain}). For noisy signals (low $\text{SNR}_{\text{est}}$), the resulting large $R$ reduces the Kalman gain $K_k$, causing the filter update (Eq.~\ref{eq:kalman_update_state}) to rely more on its internal prediction and apply stronger smoothing. Significantly, for clean signals (high $\text{SNR}_{\text{est}}$), the small $R$ increases the gain $K_k$, allowing the filter update to closely track the input $z_k$ and conservatively preserve motion details. We also provide a mathematical derivation of the steady-state filtering error and a comprehensive analysis of the SNR-driven adaptive mechanism in \textcolor{magenta}{SupMat~A-B}.

After applying the adaptive Kalman refinement to the high-frequency components ($c_k^{(j,d)}$ for $k \ge k_0$) to obtain the refined estimates $\hat{x}_{k|k}$, these filtered coefficients are combined with the original low-frequency coefficients ($c_k^{(j,d)}$ for $k < k_0$). The complete refined frequency spectrum is then transformed back into the time domain using the Inverse Discrete Cosine Transform (IDCT) for each channel $(j,d)$, yielding the final smoothed motion sequence $\widehat{\mathbf{Y}}'$. 

This adaptive design applies stronger smoothing to jittery predictions while conservatively preserving already smooth motions.
Notably, the module improves temporal consistency without compromising diversity. The justification about Kalman refinement in the frequency domain can be found in \textcolor{magenta}{SupMat~C}, and the relevant algorithm is summarized in \textcolor{magenta}{SupMat~I}.

\section{Experiments}
\subsection{Experimental Setup}
\label{sec:setup} 
We evaluate our method on the Human3.6M~\cite{ionescu2013human3} and HumanEva-I~\cite{humaneva} datasets. Performance is measured using established metrics for accuracy (ADE, FDE, MMADE, MMFDE) and diversity (APD). Our model adopts the architecture of ~\cite{SLD-HMP}, which also serves as the baseline for our experimental comparisons. We incorporate our proposed physical constraints (Sec.~\ref{sec:loss}) during training and apply the adaptive frequency-Kalman refinement (Sec.~\ref{sec:freq_kalman}) during inference. All experiments are conducted on a single NVIDIA RTX A6000 GPU. Detailed introductions to the datasets, metric definitions, hyperparameters, and network configurations are provided in \textcolor{magenta}{SupMat~D-E}.

\begin{table*}[t]
\centering
\caption{Quantitative comparison between KHMP and published state-of-the-art methods. The \textbf{bold} and \underline{underlined} values represent the best and second-best results, respectively. * indicates our re-implementation of the baseline ~\cite{SLD-HMP}, trained with KHMP's hyperparameters but without physical constraints (Sec.~\ref{sec:loss}) and Frequency-Kalman refinement (Sec.~\ref{sec:freq_kalman}).}
\label{tab:main_results}
\renewcommand\arraystretch{1.0}
\resizebox{\textwidth}{!}{
\begin{tabular}{c|c|ccccc|ccccc}
\hline
\multirow{2}{*}{\textbf{Method}} & \multirow{2}{*}{\textbf{Venue}} & \multicolumn{5}{c|}{\textbf{HumanEva-I}} & \multicolumn{5}{c}{\textbf{Human3.6M}} \\
& & APD$\uparrow$ & ADE$\downarrow$ & FDE$\downarrow$ & MMADE$\downarrow$ & MMFDE$\downarrow$ & APD$\uparrow$ & ADE$\downarrow$ & FDE$\downarrow$ & MMADE$\downarrow$ & MMFDE$\downarrow$ \\
\hline
ERD~\cite{ERD}         & ICCV 2015 & 0     & 0.382 & 0.461 & 0.521 & 0.595 & 0     & 0.722 & 0.969 & 0.776 & 0.995 \\
DeLiGAN~\cite{DeLiGAN}  & CVPR 2017 & 2.177 & 0.306 & 0.322 & 0.385 & 0.371 & 6.509 & 0.483 & 0.534 & 0.520 & 0.545 \\
BoM~\cite{BoM}         & CVPR 2018 & 2.846 & 0.271 & 0.279 & 0.373 & 0.351 & 6.265 & 0.448 & 0.533 & 0.514 & 0.544 \\
DLow~\cite{DLow}        & ECCV 2020 & 4.855 & 0.251 & 0.268 & 0.362 & 0.339 & 11.741& 0.425 & 0.518 & 0.495 & 0.531 \\
DSF~\cite{DSF}         & ICLR 2020 & 4.538 & 0.273 & 0.290 & 0.364 & 0.340 & 9.330 & 0.493 & 0.592 & 0.550 & 0.599 \\
GSPS~\cite{GSPS}        & ICCV 2021 & 5.825 & 0.233 & 0.244 & 0.343 & 0.331 & 14.757& 0.389 & 0.496 & 0.476 & 0.525 \\
MOJO~\cite{MOJO}        & CVPR 2021 & 4.181 & 0.234 & 0.260 & 0.344 & 0.339 & 12.579& 0.412 & 0.514 & 0.497 & 0.538 \\
DivSamp~\cite{DivSamp}  & ACM MM 2022 & 6.109 & 0.220 & 0.234 & 0.342 & 0.316 & 15.310 & 0.370 & 0.485 & 0.475 & 0.516 \\
STARS~\cite{STARS}      & ECCV 2022 & 6.031 & 0.217 & 0.241 & 0.328 & 0.321 & \textbf{15.884} & 0.358 & \underline{0.445} & 0.442 & 0.471 \\
MotionDiff~\cite{MotionDiff}   & AAAI 2023 & 5.931 & 0.232 & 0.236 & 0.352 & 0.320 & 15.353 & 0.411 & 0.509 & 0.508 & 0.536 \\
Belfusion~\cite{Belfusion}     & ICCV 2023 & -     & -     & -     & -     & -     & 7.602 & 0.372 & 0.474 & 0.473 & 0.507 \\
HumanMAC~\cite{HumanMAC}      & ICCV 2023 & 6.554 & 0.209 & 0.223 & 0.342 & 0.320 & 6.301 & 0.369 & 0.480 & 0.509 & 0.545 \\
TransFusion~\cite{TransFusion} & RA-L 2024 & 1.031 & 0.204 & 0.234 & 0.408 & 0.427 & 5.975 & 0.358 & 0.468 & 0.506 & 0.539 \\
CoMotion~\cite{CoMotion}       & ECCV 2024 & -     & -     & -     & -     & -     & 7.632 & 0.350 & 0.458 & 0.494 & 0.506 \\
SkeletonDiff~\cite{SkeletonDiff} & CVPR 2025 & - & - & - & - & - & 7.249 & \textbf{0.344} & 0.450 & 0.487 & 0.512 \\
MotionMap~\cite{MotionMap} & CVPR 2025 & - & - & - & - & - & 7.840 & 0.474 & 0.598 & 0.466 & 0.532 \\
SOGM~\cite{tong2025diverse} & DSP 2025 & \underline{6.761} & 0.217 & 0.217 & 0.337 & 0.322 & \underline{15.877} & 0.367 & 0.484 & 0.495 & 0.529
\\
\hline
Baseline* & -- & 6.516 & \underline{0.196} & \underline{0.211} & \underline{0.314} & \underline{0.303} & 8.217 & 0.357 & \underline{0.445} & \underline{0.438} & \underline{0.470} \\
\textbf{KHMP (Ours)} & -- & \textbf{7.481} & \textbf{0.188} & \textbf{0.204} & \textbf{0.301} & \textbf{0.291} & 9.235 & \underline{0.349} & \textbf{0.441} & \textbf{0.436} & \textbf{0.468} \\
\hline 
\end{tabular}}
\vspace{-15pt}
\end{table*}

\subsection{Comparisons with State-of-the-art Methods}
\label{sec:sota}
We evaluate KHMP against state-of-the-art methods on the HumanEva-I and Human3.6M datasets. As detailed in Table~\ref{tab:main_results}, our approach establishes a new state-of-the-art in prediction accuracy on the HumanEva-I benchmark, achieving an ADE of 0.188, FDE of 0.204, and a leading MMADE of 0.301. This performance surpasses recent methods, highlighting the effectiveness of our proposed Kalman guidance. Crucially, this top-tier accuracy is achieved without compromising diversity, a common trade-off in stochastic prediction. KHMP delivers a high and competitive APD of 7.481, significantly outperforming other accuracy-focused methods while also being more precise than many high-diversity approaches. This result validates the superiority of our frequency-domain Kalman refinement. Further justification for our adaptive strategy is provided in \textcolor{magenta}{SupMat~F}, which demonstrates its advantage over the wavelet-based manifold approach.
\vspace{-0.3in}
\section{Ablation Studies}
\label{sec:ablation}
\subsection{Effect of Frequency-Kalman Refinement}
This study isolates the contribution of our Frequency-Kalman Refinement module. We compare two settings on a model trained with our objective: (1) a standard Kalman filter with fixed noise parameters ($Q, R$ constant across all signals), and (2) our proposed adaptive Kalman filter. Table~\ref{tab:ablation_kalman} shows that while even a fixed Kalman filter offers some improvement, our adaptive strategy, which modulates $Q$ and $R$ based on the estimated SNR, achieves the most significant reduction in error metrics. This confirms that dynamically adjusting the filter's strength based on the signal's frequency characteristics is crucial for optimal performance. A more detailed analysis of SNR is illustrated in \textcolor{magenta}{SupMat~B}.

\subsection{Effect of Structured Physical Constraints}
To validate the impact of our proposed training-time losses, we incrementally add the temporal smoothness loss ($\mathcal{L}_{\text{temporal}}$) and the joint angle constraint loss ($\mathcal{L}_{\text{angle}}$) to a baseline model. The baseline model is trained with Kalman refinement but using only the standard variational objective $\mathcal{L}_{\text{VAE}}$. As shown in Table~\ref{tab:ablation_loss}, each component brings an improvement in motion quality. Adding $\mathcal{L}_{\text{temporal}}$ notably reduces jitter-related artifacts, while incorporating $\mathcal{L}_{\text{angle}}$ improves the anatomical plausibility of the generated poses. When combined, these constraints work synergistically to enhance overall realism.

\subsection{Analysis of the Full KHMP Framework}
To demonstrate the synergy between our proposed components, we compare the performance of four distinct configurations in Table~\ref{tab:ablation_full}. The results clearly indicate that both the training-time physical constraints and the Kalman refinement contribute positively. The full KHMP model, which integrates both aspects, achieves the best overall performance, demonstrating that guiding the model with physical losses during training and refining its output with our frequency-aware filter at inference time are complementary strategies. Beyond these performance improvements, we provide a broader discussion on diversity maintenance, computational efficiency, and promising future directions in \textcolor{magenta}{SupMat~J}.

\vspace{-0.1in}
\begin{table*}[t]
\centering
\caption{Comprehensive ablation studies and quantitative analysis of the KHMP framework. \textbf{(a)} Analysis of the full framework components. \textbf{(b)} Ablation on structured physical losses. \textbf{(c)} Comparison of Frequency-Kalman refinement strategies. \textbf{(d)} Fixed Frequency Suppression~\cite{wu2024frequency} vs. our Adaptive Frequency-Kalman Refinement on HumanEva-I. \textbf{(e)} Quantitative evaluation of jitter reduction.}
\label{tab:combined_ablations}
\vspace{-15pt} 
\renewcommand{\arraystretch}{0.9} 

\begin{minipage}[t]{0.44\linewidth}
    \vspace{0pt} 
    
    \begin{subtable}[t]{\linewidth}
        \centering
        \resizebox{\linewidth}{!}{
        \begin{tabular}{lcccc} 
        \toprule
        \textbf{Method} & \textbf{Phys. Loss} & \textbf{Kalman} & ADE$\downarrow$ & FDE$\downarrow$ \\
        \midrule
        Baseline & \ding{55} & \ding{55} & 0.196 & 0.211 \\
        w/o Phys. Loss & \ding{55} & \ding{51} & 0.193 & 0.208 \\
        w/o Kalman & \ding{51} & \ding{55} & 0.194 & 0.206 \\
        \rowcolor{gray!20} \textbf{KHMP (Full)} & \ding{51} & \ding{51} & \textbf{0.188} & \textbf{0.204} \\
        \bottomrule
        \end{tabular}}
        \caption{Full KHMP framework.}
        \label{tab:ablation_full}
    \end{subtable}
    \begin{subtable}[t]{\linewidth}
        \centering
        \renewcommand\arraystretch{0.3}
        \resizebox{\linewidth}{!}{
        \begin{tabular}{lcccc} 
        \toprule
        \textbf{Method} & $\mathcal{L}_{\text{temporal}}$ & $\mathcal{L}_{\text{angle}}$ & ADE$\downarrow$ & FDE$\downarrow$ \\
        \midrule
        Baseline & \ding{55} & \ding{55} & 0.193 & 0.208 \\
        w/o $\mathcal{L}_{\text{angle}}$ & \ding{51} & \ding{55} & 0.191 & 0.208 \\
        w/o $\mathcal{L}_{\text{temp}}$ & \ding{55} & \ding{51} & 0.190 & 0.206 \\
        \rowcolor{gray!20} \textbf{KHMP} & \ding{51} & \ding{51} & \textbf{0.188} & \textbf{0.204} \\
        \bottomrule
        \end{tabular}}
        \caption{Structured physical losses.}
        \label{tab:ablation_loss}
    \end{subtable}
    \begin{subtable}[t]{\linewidth}
        \centering
        \resizebox{\linewidth}{!}{
        \begin{tabular}{lccc}
        \toprule
        \textbf{Refinement Strategy} & ADE$\downarrow$ & FDE$\downarrow$ & APD$\uparrow$ \\
        \midrule
        Fixed Kalman & 0.194 & 0.209 & 6.208 \\
        \rowcolor{gray!20} \textbf{Adaptive (Ours)} & \textbf{0.188} & \textbf{0.204} & \textbf{7.481} \\
        \bottomrule
        \end{tabular}}
        \caption{Kalman refinement.}
        \label{tab:ablation_kalman}
    \end{subtable}

\end{minipage}%
\hfill
\begin{minipage}[t]{0.54\linewidth}
    \vspace{0pt} 

    \begin{subtable}[t]{\linewidth}
        \centering
        \resizebox{\linewidth}{!}{
        \begin{tabular}{lccccc}
        \toprule
        \textbf{Method} & APD$\uparrow$ & ADE$\downarrow$ & FDE$\downarrow$ & MMADE$\downarrow$ & MMFDE$\downarrow$ \\
        \midrule
        Fix Supp. ($\gamma=0.1$) & 4.892 & 0.203 & 0.221 & 0.328 & 0.316 \\
        Fix Supp. ($\gamma=0.2$) & 5.341 & 0.200 & 0.218 & 0.324 & 0.313 \\
        Fix Supp. ($\gamma=0.3$) & \underline{5.921} & \underline{0.198} & \underline{0.213} & 0.318 & \underline{0.307} \\
        Fix Supp. ($\gamma=0.4$) & 5.834 & 0.203 & 0.214 & \underline{0.317} & 0.308 \\
        Fix Supp. ($\gamma=0.5$) & 5.627 & 0.202 & 0.217 & 0.321 & 0.310 \\
        Fix Supp. ($\gamma=0.6$) & 5.748 & 0.201 & \underline{0.213} & 0.320 & 0.309 \\
        Fix Supp. ($\gamma=0.7$) & 5.429 & 0.205 & 0.217 & 0.323 & 0.312 \\
        Fix Supp. ($\gamma=0.8$) & 5.016 & 0.207 & 0.220 & 0.326 & 0.315 \\
        Fix Supp. ($\gamma=0.9$) & 4.573 & 0.210 & 0.223 & 0.330 & 0.318 \\
        \midrule
        \rowcolor{gray!20} \textbf{KHMP (Ours)} & \textbf{7.481} & \textbf{0.188} & \textbf{0.204} & \textbf{0.301} & \textbf{0.291} \\
        \bottomrule
        \end{tabular}}
        \caption{Fixed suppression vs. Ours.}
        \label{tab:fixed_vs_adaptive_ablation}
    \end{subtable}
    \vspace{15pt}
    \begin{subtable}[t]{\linewidth}
        \centering
        \resizebox{\linewidth}{!}{
        \setlength{\tabcolsep}{3pt} 
        \begin{tabular}{lccc @{\hspace{8pt}} lccc}
        \toprule
        \textbf{Body Part} & \textbf{Base} & \textbf{Ours} & \textbf{Reduction} & \textbf{Body Part} & \textbf{Base} & \textbf{Ours} & \textbf{Reduction} \\
        \midrule
        Pelvis & 0.20 & \textbf{0.17} & 15.0\% $\downarrow$ & Neck & 0.30 & \textbf{0.26} & 13.3\% $\downarrow$ \\
        Spine & 0.25 & \textbf{0.21} & 16.0\% $\downarrow$ & Head & 0.35 & \textbf{0.30} & 14.3\% $\downarrow$ \\
        Hip & 0.22 & \textbf{0.19} & 13.6\% $\downarrow$ & Shoulder & 0.32 & \textbf{0.27} & 15.6\% $\downarrow$ \\
        Thigh & 0.38 & \textbf{0.30} & 21.1\% $\downarrow$ & Upper Arm & 0.41 & \textbf{0.24} & 41.5\% $\downarrow$ \\
        Shin & 0.48 & \textbf{0.29} & 39.6\% $\downarrow$ & Forearm & 0.55 & \textbf{0.40} & 27.3\% $\downarrow$ \\
        Ankle & 0.52 & \textbf{0.31} & 40.4\% $\downarrow$ & Wrist & 0.63 & \textbf{0.38} & 39.7\% $\downarrow$ \\
        \midrule
        \rowcolor{gray!20} \textbf{Average} & 0.38 & \textbf{0.28} & \textbf{28.0\%} $\downarrow$ & \multicolumn{4}{c}{\cellcolor{gray!20}\textit{(Calculated across all parts)}} \\
        \bottomrule
        \end{tabular}}
        \caption{Quantitative jitter reduction ($\times 10^{-3}$).}
        \label{tab:jitter_metrics}
    \end{subtable}

\end{minipage}
\vspace{-25pt}
\end{table*}

\begin{figure*}[t]
    \centering
    \begin{subfigure}[b]{0.19\textwidth}
        \includegraphics[width=\linewidth]{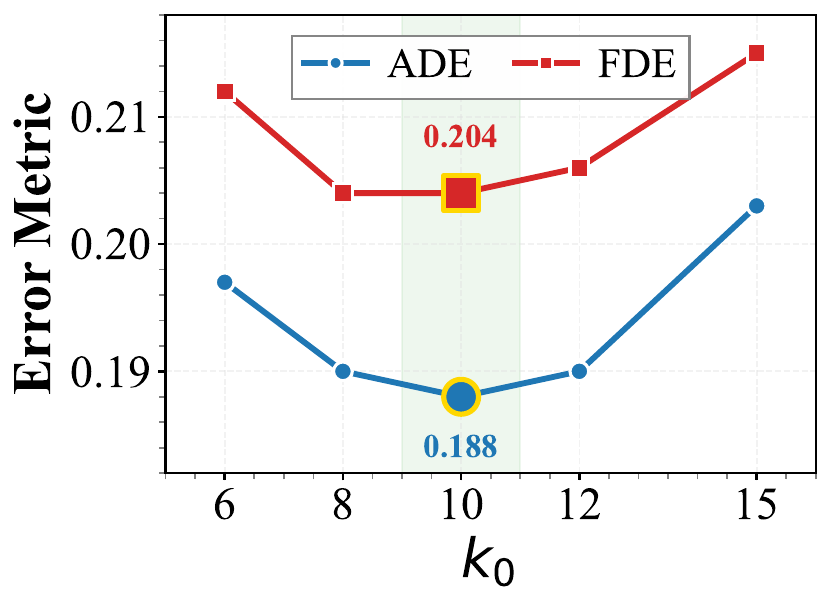}
        \caption{Impact of $k_0$.}
        \label{fig:ablation_k0}
    \end{subfigure}
    \hfill 
    \begin{subfigure}[b]{0.19\textwidth}
        \includegraphics[width=\linewidth]{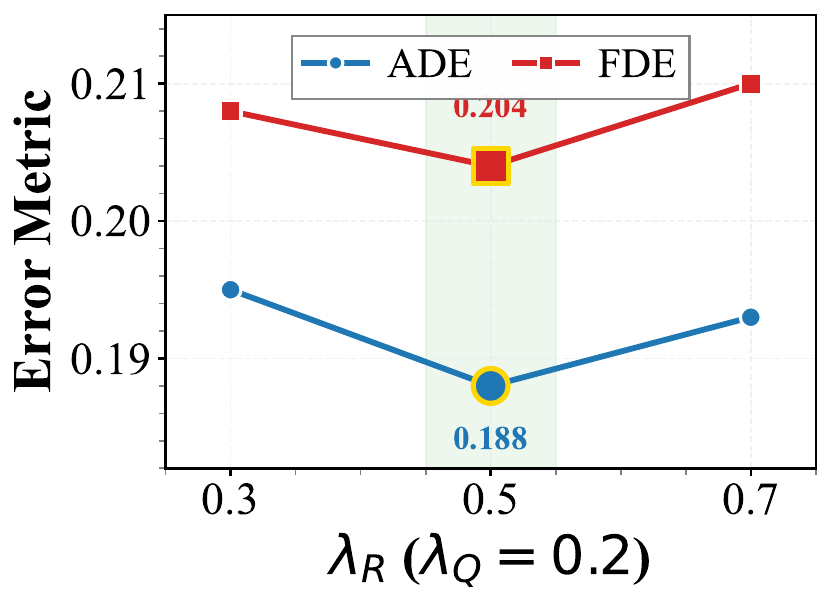}
        \caption{Impact of $\lambda_R$.}
        \label{fig:ablation_lambda_r}
    \end{subfigure}
    \hfill 
    \begin{subfigure}[b]{0.19\textwidth}
        \includegraphics[width=\linewidth]{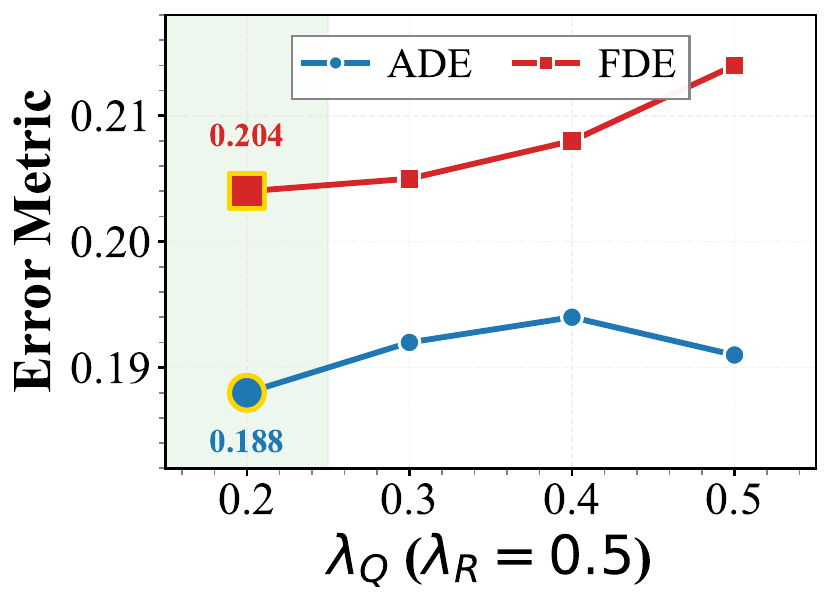}
        \caption{Impact of $\lambda_Q$.}
        \label{fig:ablation_lambda_q}
    \end{subfigure}
    \hfill 
    \begin{subfigure}[b]{0.19\textwidth}
        \includegraphics[width=\linewidth]{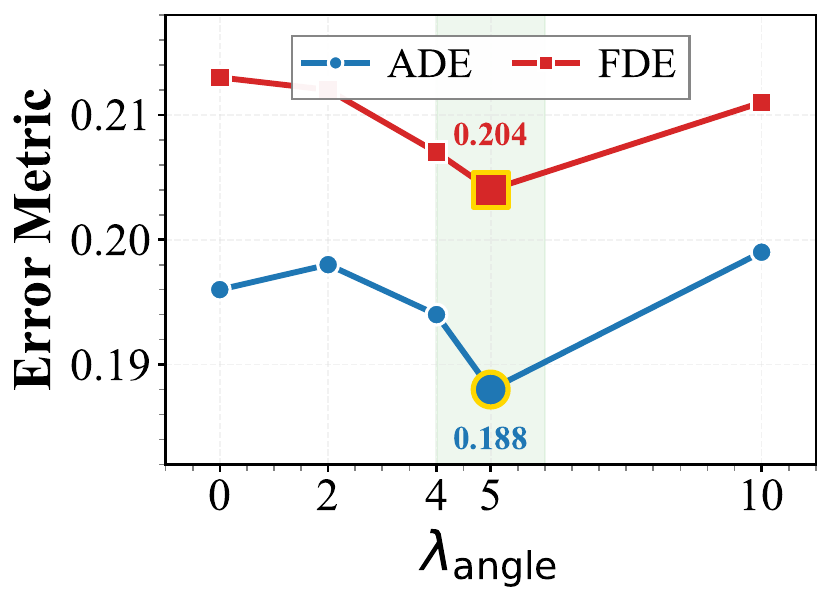}
        \caption{Impact of $\lambda_{\text{angle}}$.}
        \label{fig:ablation_lambda_angle}
    \end{subfigure}
    \hfill 
    \begin{subfigure}[b]{0.19\textwidth}
        \includegraphics[width=\linewidth]{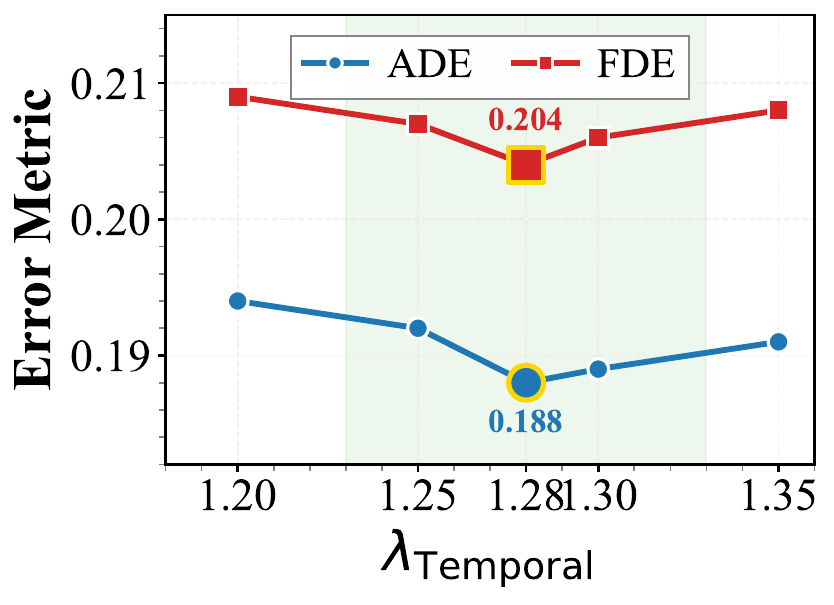}
        \caption{Impact of $\lambda_{\text{Temporal}}$.}
        \label{fig:ablation_lambda_temporal}
    \end{subfigure}
    \vspace{-5pt}
    \caption{Sensitivity analysis of hyperparameters: $k_0$, $\lambda_Q$, $\lambda_R$, $\lambda_{\text{angle}}$, and $\lambda_{\text{temporal}}$.}
    \label{fig:ablation_all}
    \vspace{-20pt}
\end{figure*}

\subsection{Analysis of Hyperparameters}
We analyze the sensitivity of key hyperparameters in our method, with results visualized in Figure~\ref{fig:ablation_all}. First, we vary the DCT threshold $k_0$, which separates frequency components. Plot (a) shows a clear performance optimum at $k_0=10$. Next, we study the SNR sensitivity parameters $\lambda_Q$ and $\lambda_R$. Plot (b) fixes $\lambda_Q=0.2$ and varies $\lambda_R$, finding the best performance at 0.5. Plot (c) then fixes $\lambda_R=0.5$ and varies $\lambda_Q$, confirming 0.2 as the optimal choice. We further examine the angle constraint weight $\lambda_{\text{angle}}$ in plot (d), achieving optimal results at 5, and the temporal smoothness weight $\lambda_{\text{Temporal}}$ in plot (e), with the best performance at 1.28. The results indicate that our default settings provide a good balance and that performance remains stable across a range of values, demonstrating the robustness of our adaptive strategy for human motion prediction.

\begin{figure*}[!]
    \centering
    \begin{subfigure}[b]{0.49\linewidth} 
        \centering
        \includegraphics[width=\linewidth]{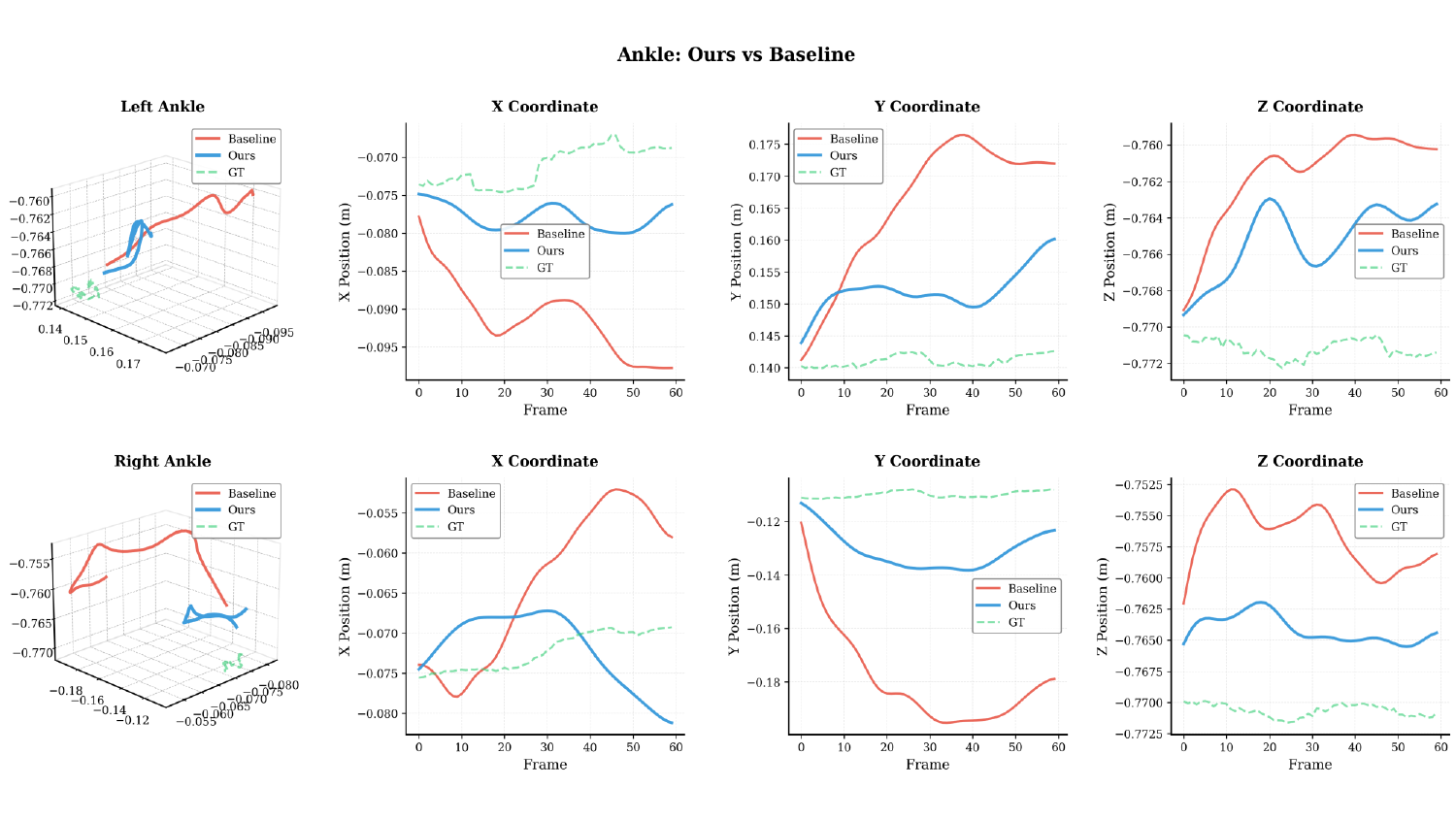}
        \caption{\textit{Gesturing} - Ankle: Ours vs Baseline}
        \label{fig:gesturing_ankle} 
    \end{subfigure}
    \hfill 
    \begin{subfigure}[b]{0.49\linewidth} 
        \centering
        \includegraphics[width=\linewidth]{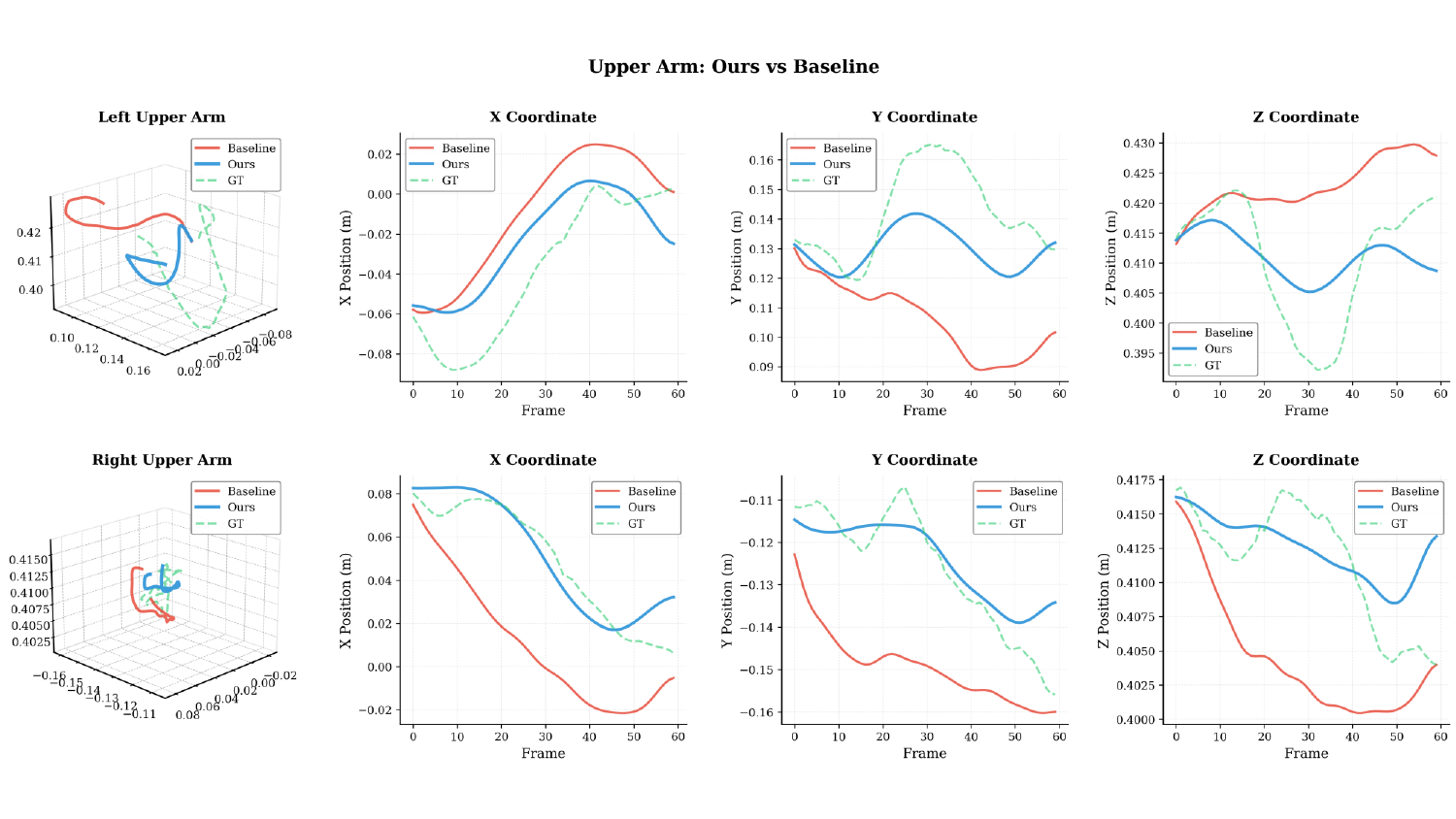}
        \caption{\textit{Gesturing} - Upper Arm: Ours vs Baseline}
        \label{fig:gesturing_upper_arm} 
    \end{subfigure}
    
    
    \begin{subfigure}[b]{0.49\linewidth} 
        \centering
        \includegraphics[width=\linewidth]{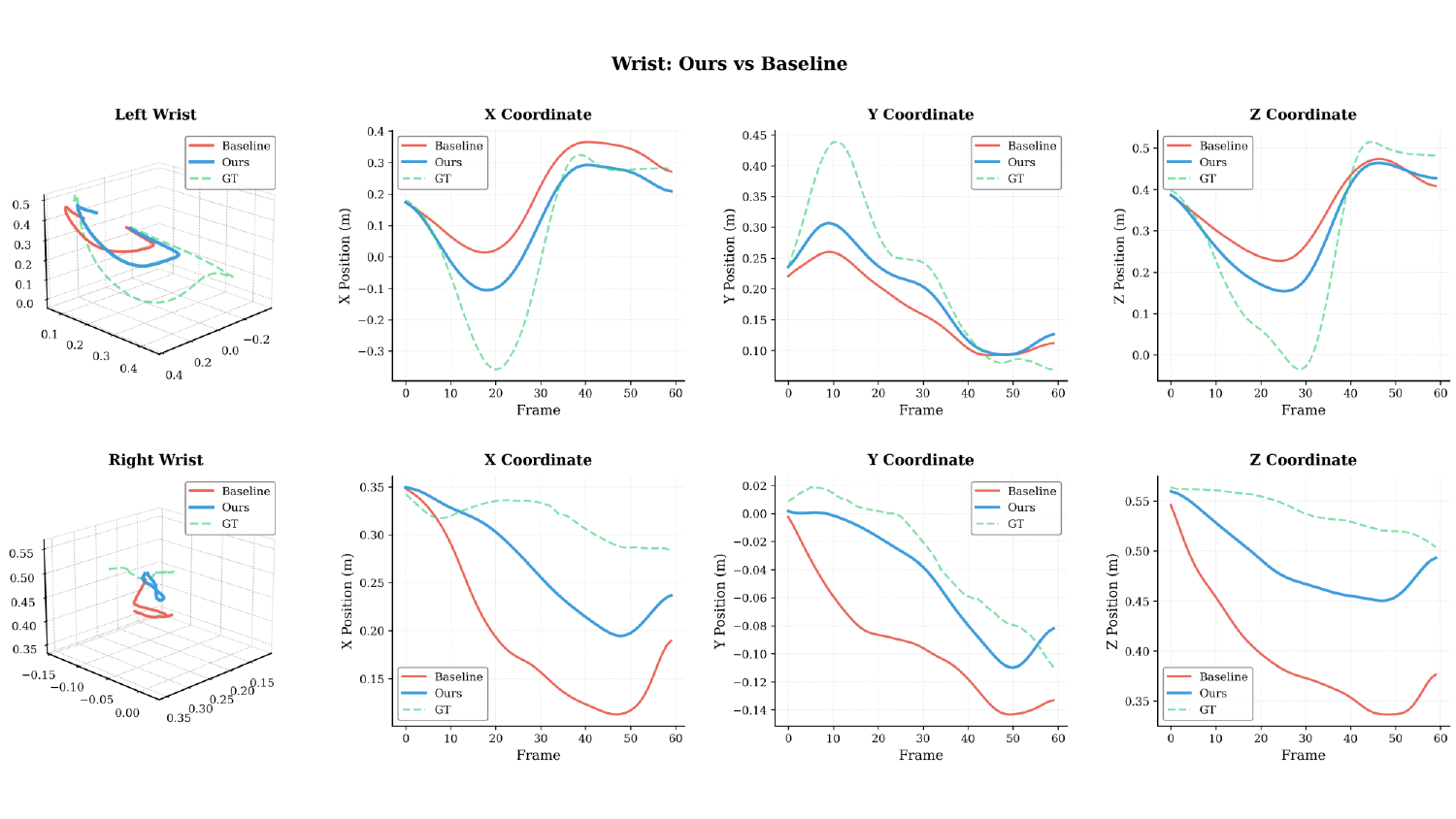} 
        \caption{\textit{Gesturing} - Wrist: Ours vs Baseline} 
        \label{fig:gesturing_wrist} 
    \end{subfigure}
    \hfill 
    \begin{subfigure}[b]{0.49\linewidth} 
        \centering
        \includegraphics[width=\linewidth]{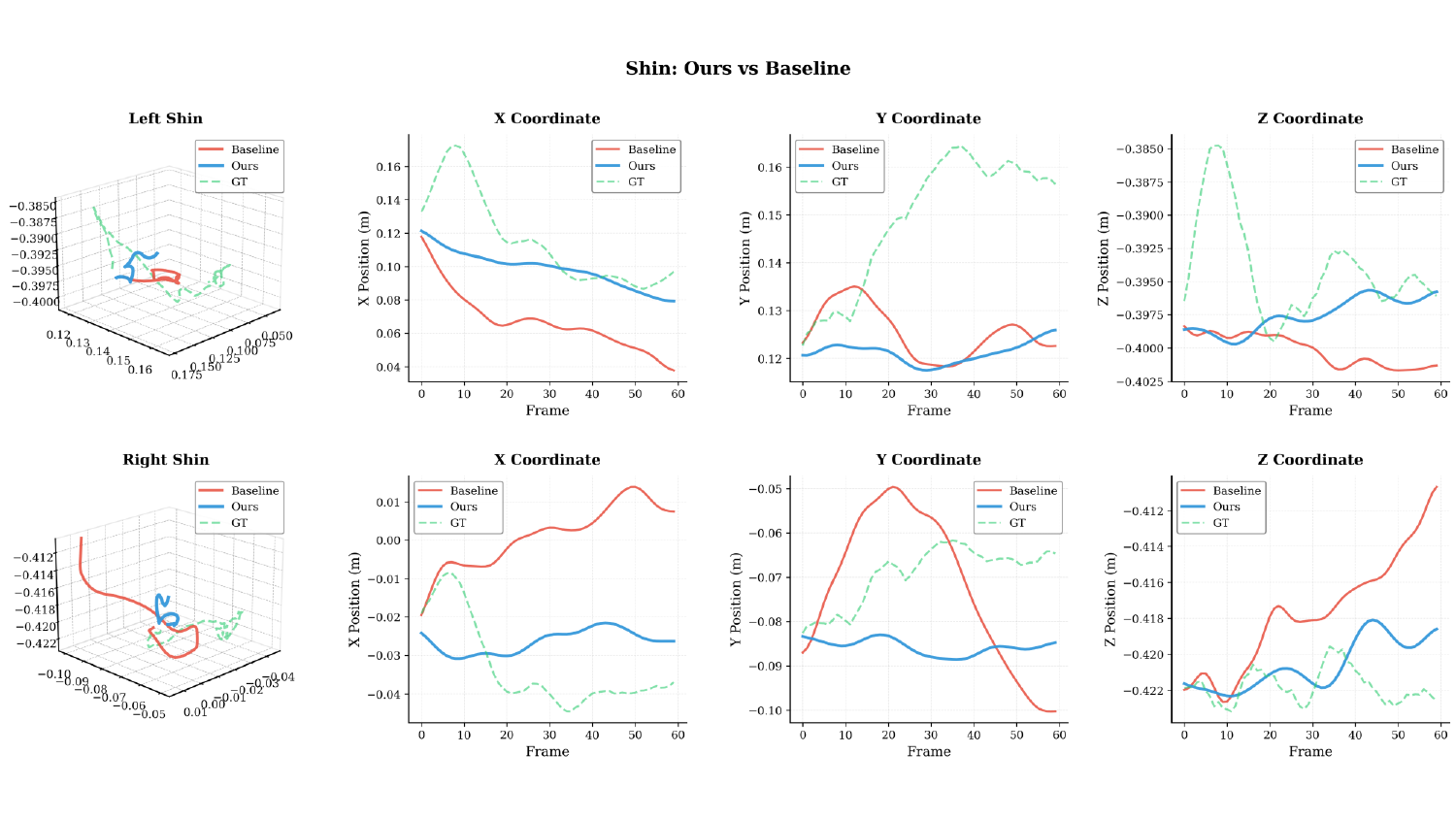} 
        \caption{\textit{Gesturing} - Shin: Ours vs Baseline} 
        \label{fig:gesturing_shin} 
    \end{subfigure}
    
    
    \begin{subfigure}[b]{0.49\linewidth} 
        \centering
        \includegraphics[width=\linewidth]{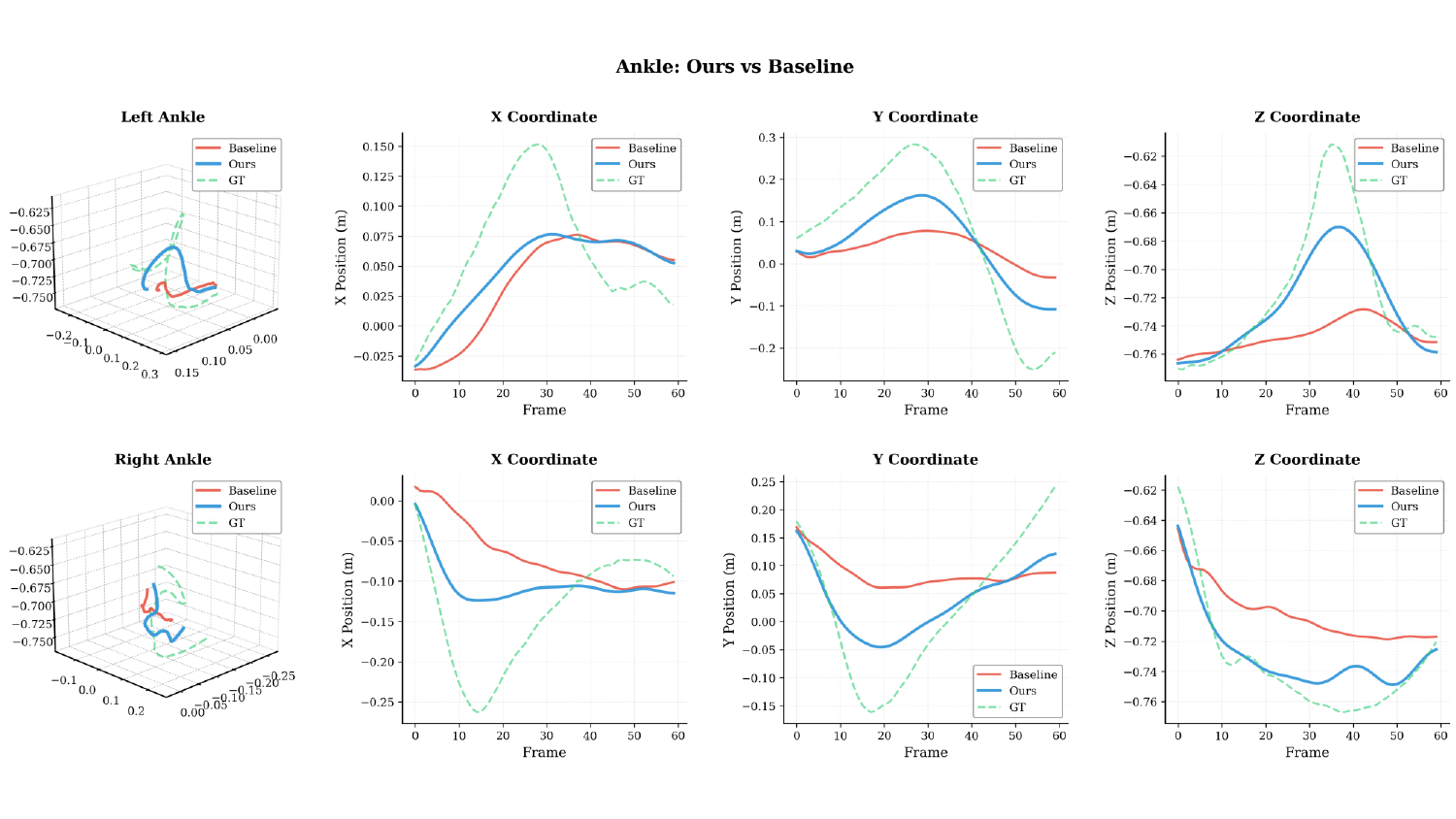}
        \caption{\textit{Walking} - Ankle: Ours vs Baseline}
        \label{fig:walking_ankle} 
    \end{subfigure}
    \hfill 
    \begin{subfigure}[b]{0.49\linewidth} 
        \centering
        \includegraphics[width=\linewidth]{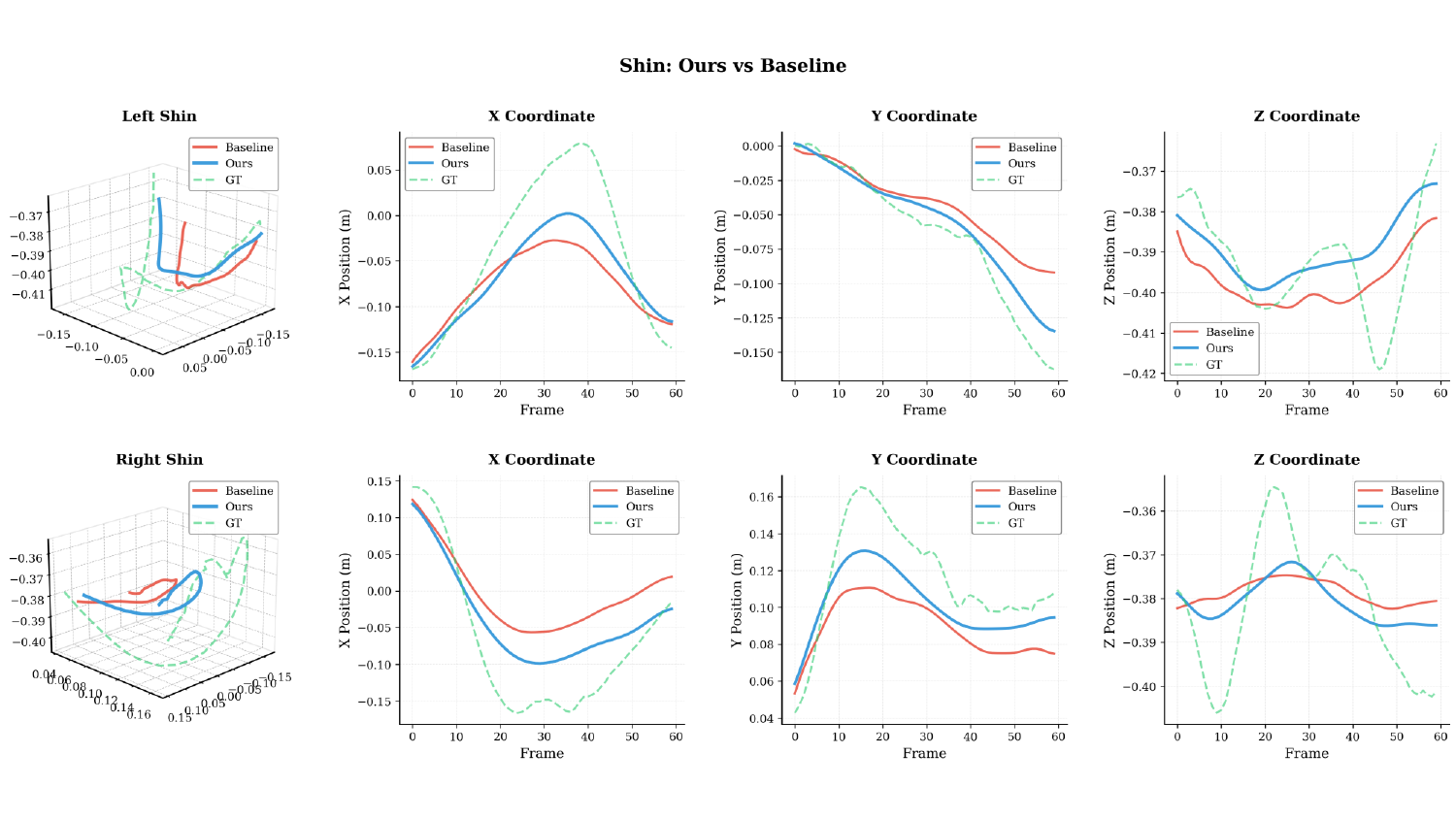}
        \caption{\textit{Walking} - Shin: Ours vs Baseline}
        \label{fig:walking_shin} 
    \end{subfigure}
    
    
    \begin{subfigure}[b]{0.49\linewidth} 
        \centering
        \includegraphics[width=\linewidth]{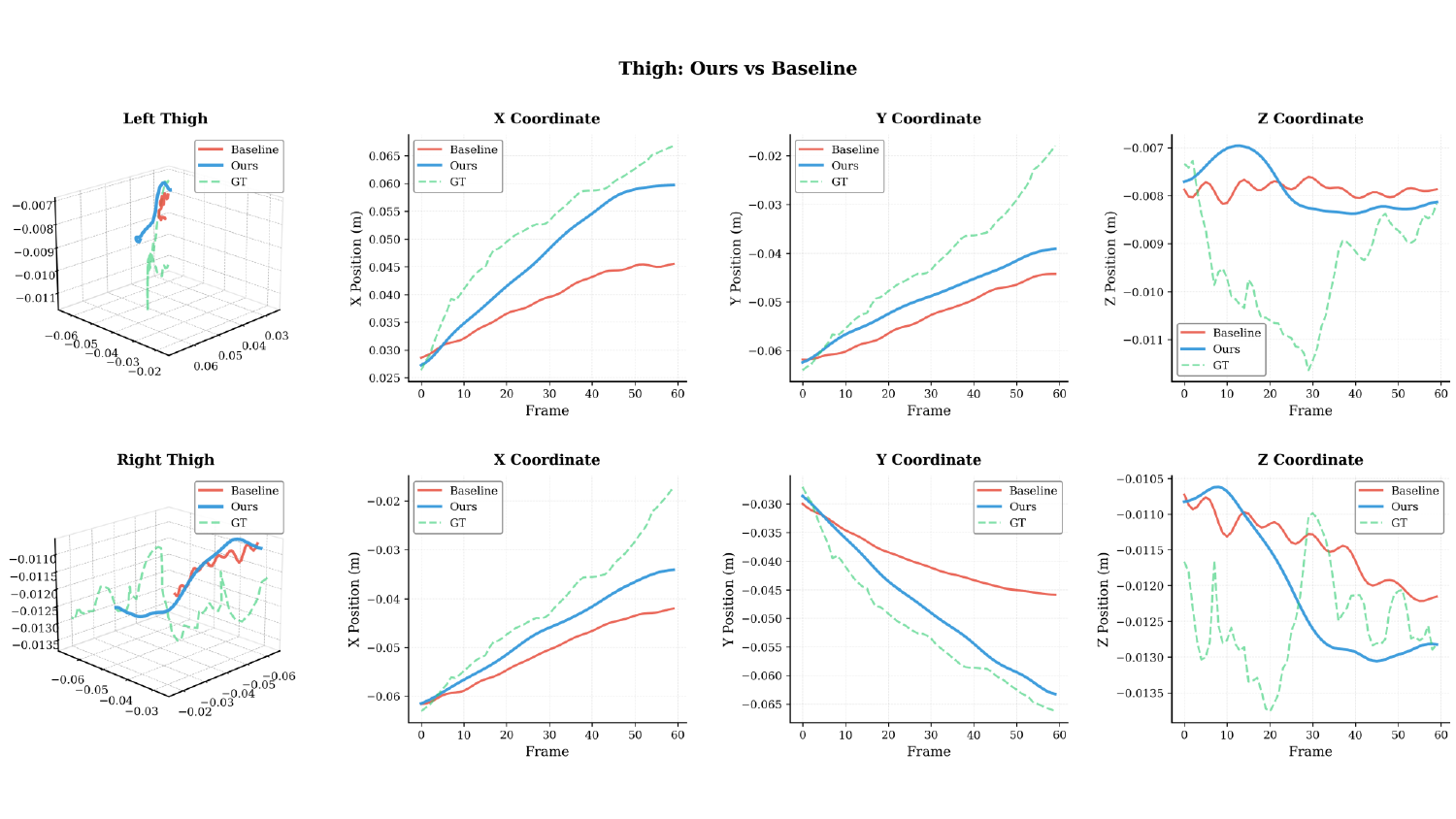} 
        \caption{\textit{Walking} - Thigh: Ours vs Baseline} 
        \label{fig:walking_thigh} 
    \end{subfigure}
    \hfill 
    \begin{subfigure}[b]{0.49\linewidth} 
        \centering
        \includegraphics[width=\linewidth]{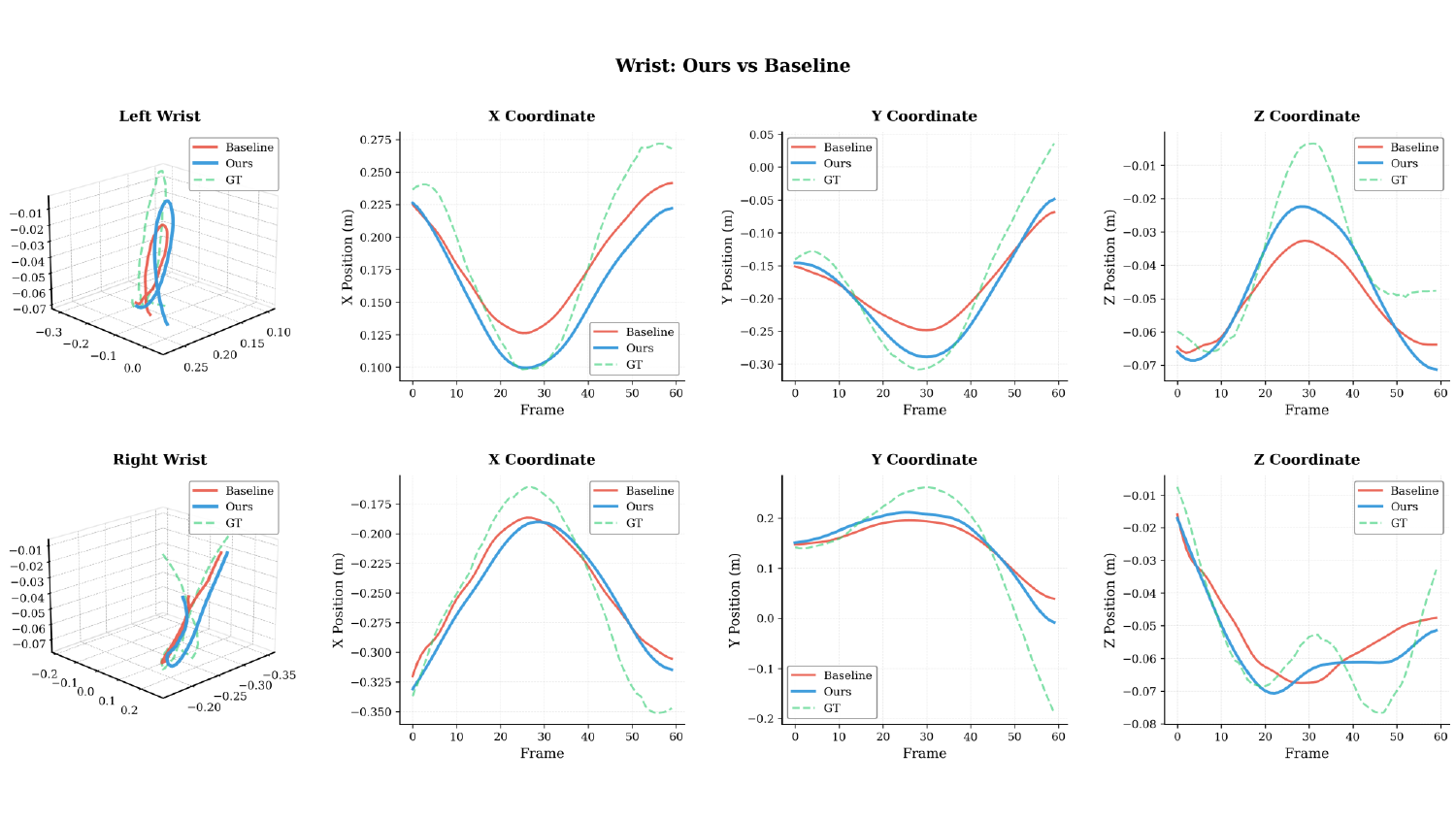} 
        \caption{\textit{Walking} - Wrist: Ours vs Baseline} 
        \label{fig:walking_wrist} 
    \end{subfigure}

\caption{
    Visual comparison of joint trajectories demonstrating KHMP's refinement over the baseline across two distinct actions. 
    The top two rows (\textbf{a}-\textbf{d}) present examples from the \textit{Gesturing} action (Ankle, Upper Arm, Wrist, and Shin). 
    The bottom two rows (\textbf{e}-\textbf{h}) present examples from the \textit{Walking} action (Ankle, Shin, Thigh, and Wrist). 
    Within each subfigure, the first column displays the 3D trajectory, while the subsequent columns show the X, Y, and Z coordinates over time. 
    Across all examples, the \textbf{\textcolor{red}{baseline}} prediction (red line) exhibits noticeable high-frequency jitter, whereas \textbf{\textcolor{blue}{KHMP}} (blue line) significantly smooths the trajectory, closely adhering to the \textbf{\textcolor{green}{ground truth}} (green dashed line) and demonstrating effective refinement. \underline{Best viewed by zooming in.}
}
    \label{fig:combined_trajectory_comparison} 
    \vspace{-0.2in}
\end{figure*}


\section{Comprehensive Analysis}
\subsection{Jitter Reduction Analysis}
\label{sec:jitter_quant}

To evaluate our refinement module's ability to suppress high-frequency jitter, we compute the third derivative of position (jerk) as a standard smoothness metric. As reported in Table~\ref{tab:jitter_metrics}, KHMP achieves a 28.0\% average jitter reduction across all body parts. Crucially, this drastic smoothing coincides with improved overall prediction accuracy (ADE: $0.196\!\to\!0.188$, Table~\ref{tab:ablation_full}), proving that our adaptive Frequency-Kalman refinement effectively removes jittery noise rather than useful motion details.

Complementing this quantitative evidence, Figure~\ref{fig:combined_trajectory_comparison} visually compares trajectories for selected body parts whose complex movements can reveal prediction jitter across the \textit{Gesturing} (Fig.~\ref{fig:gesturing_ankle}-\ref{fig:gesturing_shin}) and \textit{Walking} (Fig.~\ref{fig:walking_ankle}-\ref{fig:walking_wrist}) actions. The baseline predictions (rendered in red) exhibit rapid fluctuations and deviations from the smooth ground truth path (green, dashed). In contrast, our refined KHMP predictions (blue) exhibit improved temporal smoothness and closer adherence to the ground-truth trajectory in both 3D space and across individual coordinates. This visual evidence confirms that our adaptive method attenuates jitter while preserving motion dynamics, leading to higher temporal fidelity.

\subsection{Comparison to Fixed Frequency-Domain Smoothing}
\label{comparison_fixed_smoothing}
To justify our adaptive strategy, we compare it against a non-adaptive frequency smoothing approach inspired by FreqMixFormer~\cite{wu2024frequency}. After applying DCT to the raw predictions $\widehat{\mathbf{Y}}$ to obtain coefficients $c_k^{(j,d)}$, we multiply the high-frequency components ($k \ge k_0$) by a \textit{fixed} suppression factor $\gamma$, yielding $c'_{\text{high}, k} = \gamma \cdot c_k^{(j,d)}$. These are combined with unmodified low-frequency components ($k < k_0$) and reconstructed via IDCT. This uniformly suppresses high frequencies, completely ignoring sample-specific noise levels. Conversely, our KHMP adaptively scales Kalman parameters based on estimated SNR to recursively smooth the high-frequency spectrum. As Table~\ref{tab:fixed_vs_adaptive_ablation} demonstrates, KHMP significantly outperforms the fixed suppression method across all tested factors $\gamma \in [0.1, 0.9]$. This highlights a critical limitation of non-adaptive approaches: uniform smoothing cannot effectively adjust to the varying jitter levels inherent in diverse stochastic predictions.

\subsection{Visual Comparisons}
\label{sec:qualitative}
\begin{figure*}[h]
    \centering
    \begin{subfigure}[b]{0.49\textwidth} 
        \centering
        \includegraphics[width=\linewidth, height=0.49\textheight, keepaspectratio]{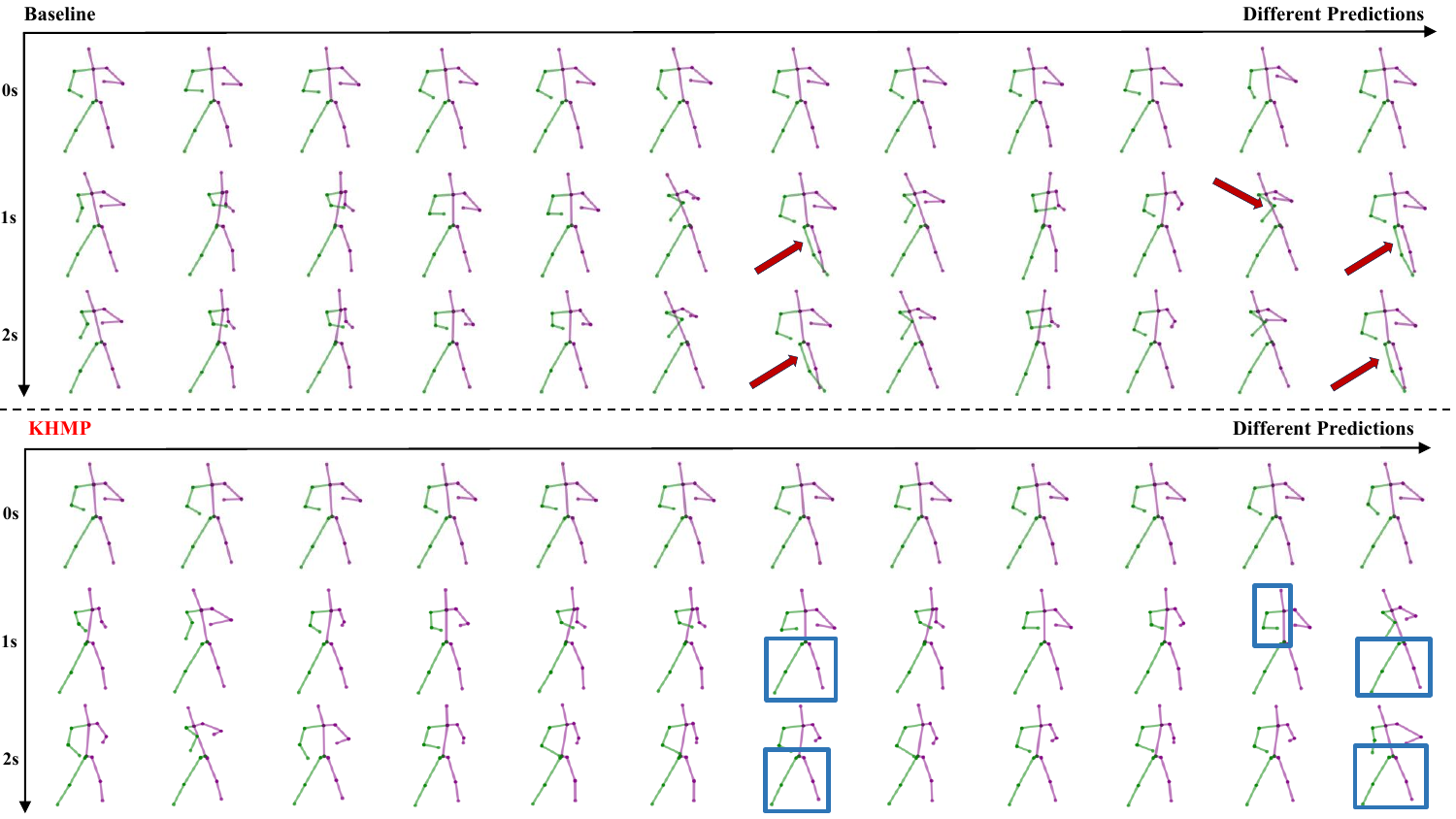}
        \caption{Qualitative comparison on \textit{boxing} action.}
        \label{fig:fig4.1_sub}
    \end{subfigure}
    \hfill 
    \begin{subfigure}[b]{0.49\textwidth} 
        \centering
        \includegraphics[width=\linewidth, height=0.49\textheight, keepaspectratio]{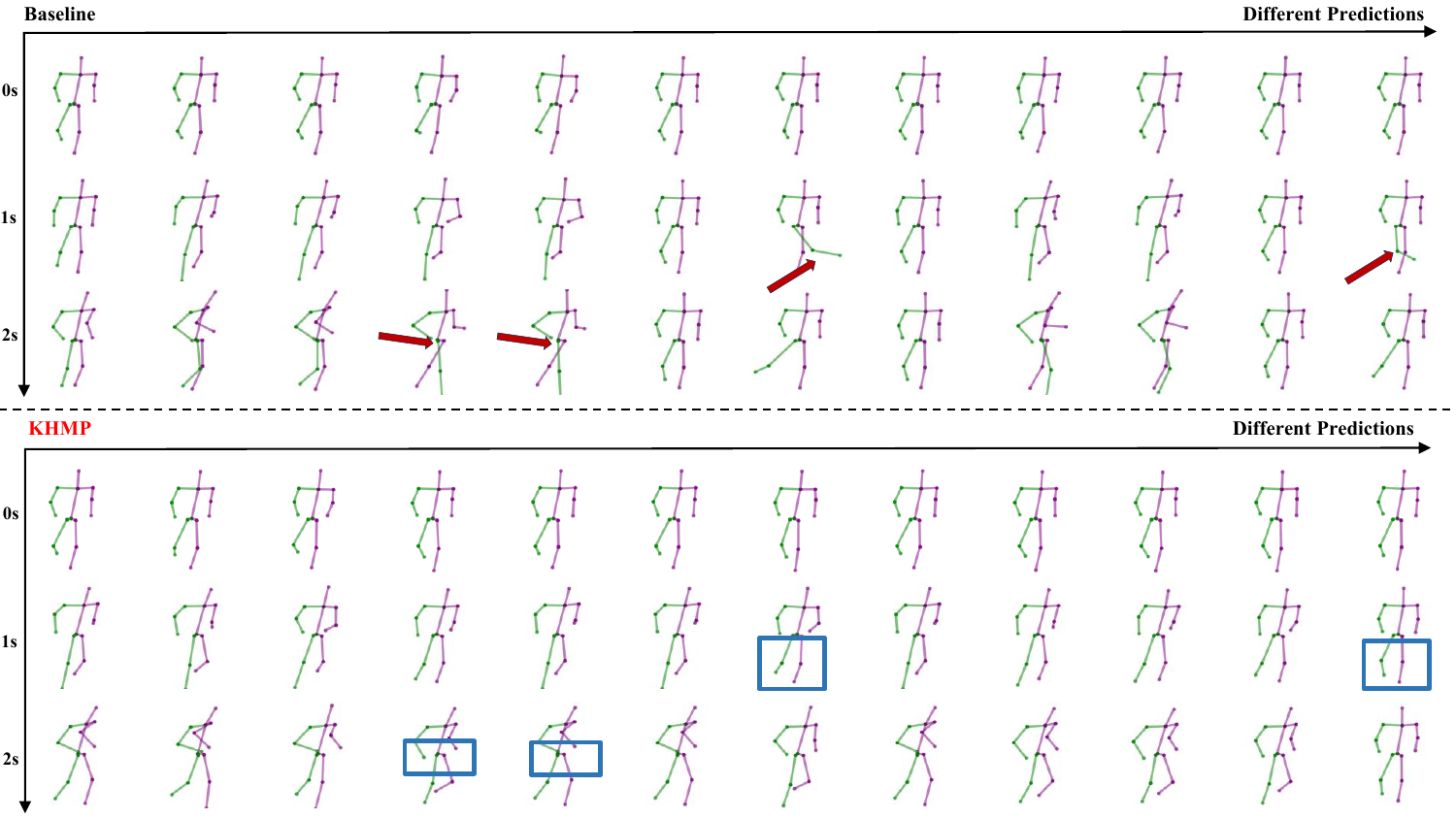}
        \caption{Qualitative comparison on \textit{jogging} action.}
        \label{fig:fig4.2_sub}
    \end{subfigure}
    \vspace{0em} 
    \begin{subfigure}[b]{0.49\textwidth} 
        \centering
        \includegraphics[width=\linewidth, height=0.49\textheight, keepaspectratio]{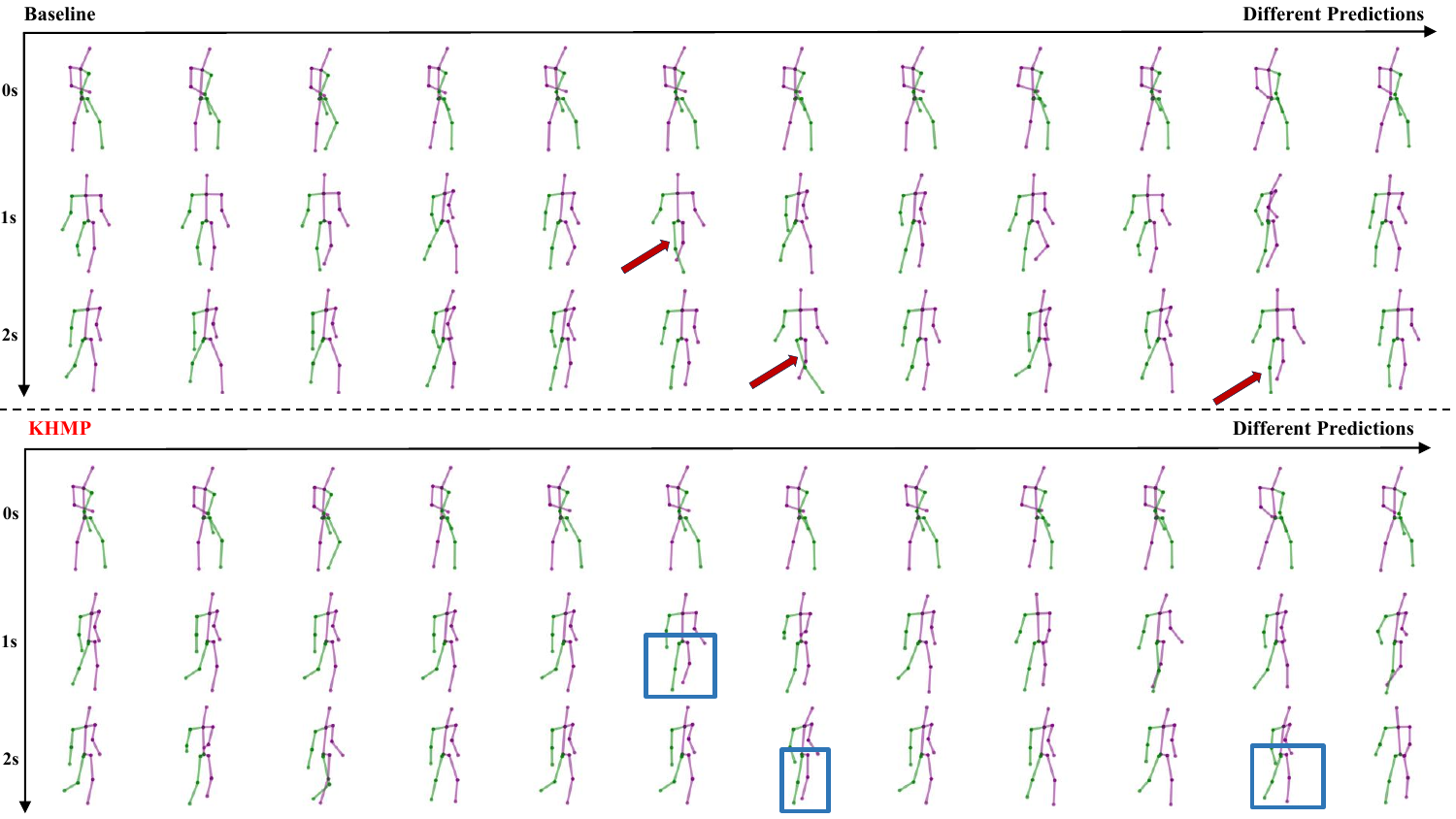} 
        \caption{Qualitative comparison on \textit{walking} action.}
        \label{fig:fig4.3_sub} 
    \end{subfigure}
    \hfill 
    \begin{subfigure}[b]{0.49\textwidth} 
        \centering
        \includegraphics[width=\linewidth, height=0.49\textheight, keepaspectratio]{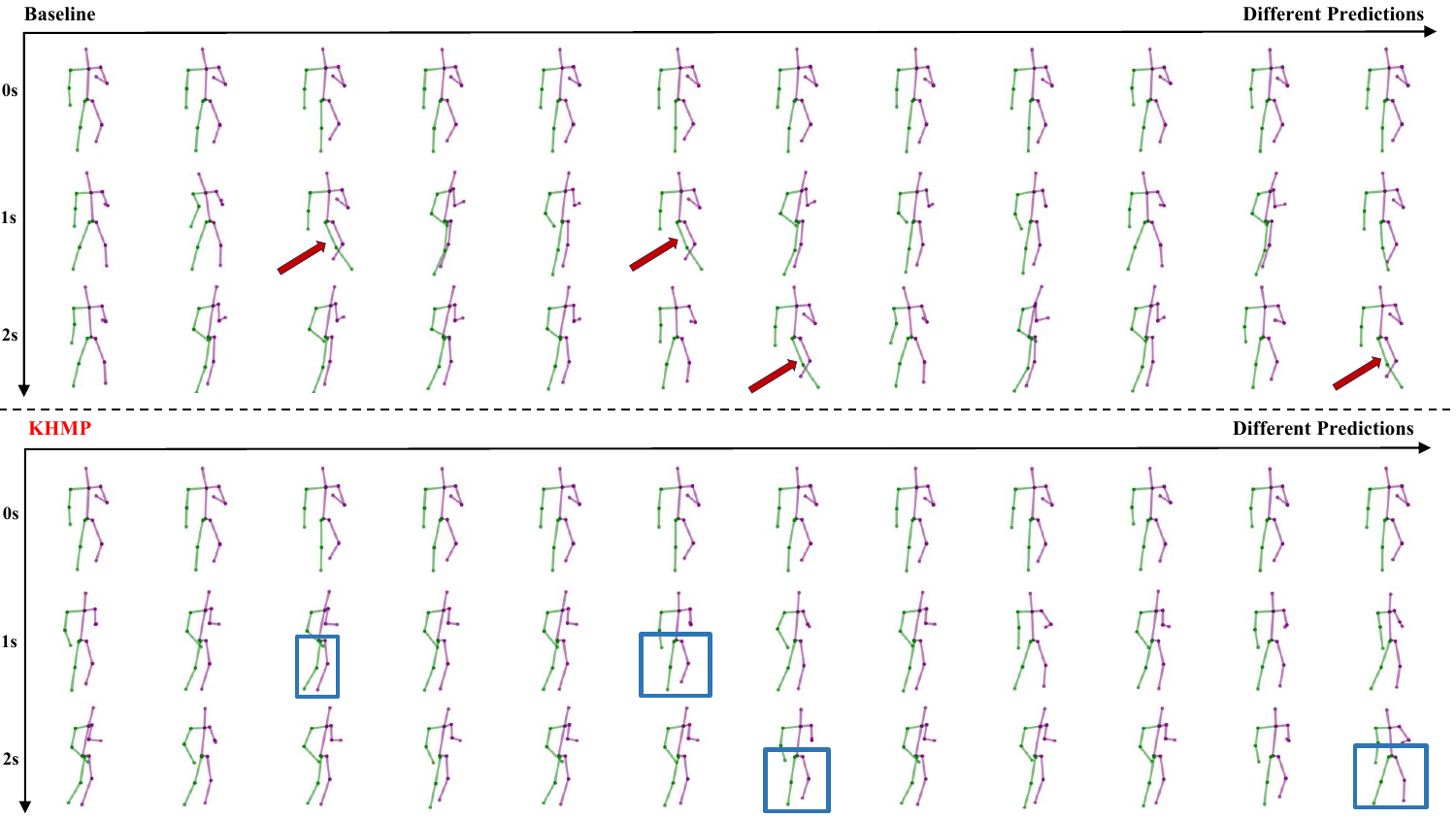} 
        \caption{Qualitative comparison on \textit{gesturing} action.} 
        \label{fig:fig4.4_sub} 
    \end{subfigure}

    \caption{
        Visual comparison highlighting KHMP's enhanced prediction quality over the baseline across various actions (\textit{Boxing}, \textit{Jogging}, \textit{Walking}, and \textit{Gesturing}). The figure showcases KHMP's ability to correct physical implausibility often present in raw generative model outputs. In the examples shown, \textcolor{red}{red arrows} indicate issues in baseline predictions, while \textcolor{blue}{blue boxes} highlight the corresponding smoother and more coherent KHMP results. \underline{Best viewed by zooming in. We also provide video demos in the \textcolor{magenta}{SupMat}.}
    }
    \label{fig:qualitative_comparison} 
    \vspace{-25pt}
\end{figure*}

To provide direct evidence of motion quality, we present qualitative comparisons of the generated sequences across various actions in Figure~\ref{fig:qualitative_comparison}. Side-by-side comparisons reveal that baseline predictions exhibit noticeable jitter or physical implausibility in some cases, particularly visible at later timesteps. For instance, in the \textit{boxing} example (Fig.~\ref{fig:fig4.1_sub}), the baseline's left arm becomes uncoordinated, twisting unnaturally, while KHMP corrects this inconsistent limb positioning. Similarly, examples for \textit{jogging} (Fig.~\ref{fig:fig4.2_sub}), \textit{walking} (Fig.~\ref{fig:fig4.3_sub}), and \textit{gesturing} (Fig.~\ref{fig:fig4.4_sub}) further illustrate these issues in the baseline. In contrast, KHMP consistently produces temporally smoother and more natural motions that remain coherent over the prediction horizon. This validates that our refinement module enhances prediction fidelity without sacrificing the generative model's diversity, generating futures that are both distinct and realistic. More visual comparisons are provided in \textcolor{magenta}{SupMat G}.

\section{Conclusion}
\label{sec:conclusion}
In this paper, we propose KHMP, a dual-pronged framework that enhances the fidelity of human motion prediction. It incorporates structured physical constraints (temporal smoothness, joint angle limits) into the training and introduces a novel, adaptive Kalman filter in the frequency domain to suppress jitter. Our experiments show that this combination significantly improves prediction accuracy and temporal coherence, achieving state-of-the-art results. By suppressing jitter and anatomical impossibilities, KHMP generates futures that are precise and plausible.

\bibliographystyle{splncs04}
\bibliography{main}

@String(CVPR  = {IEEE Conf. Comput. Vis. Pattern Recog.})

@String(AAAI  = {AAAI})

@String(CVPR  = {CVPR})

@inproceedings{ERD,
  title={Recurrent network models for human dynamics},
  author={Fragkiadaki, Katerina and Levine, Sergey and Felsen, Panna and Malik, Jitendra},
  booktitle={Proceedings of the IEEE international conference on computer vision},
  pages={4346--4354},
  year={2015}
}

@article{acLSTM,
  title={Auto-conditioned recurrent networks for extended complex human motion synthesis},
  author={Li, Zimo and Zhou, Yi and Xiao, Shuangjiu and He, Chong and Huang, Zeng and Li, Hao},
  journal={arXiv preprint arXiv:1707.05363},
  year={2017}
}

@inproceedings{DeLiGAN,
  title={Deligan: Generative adversarial networks for diverse and limited data},
  author={Gurumurthy, Swaminathan and Kiran Sarvadevabhatla, Ravi and Venkatesh Babu, R},
  booktitle={Proceedings of the IEEE conference on computer vision and pattern recognition},
  pages={166--174},
  year={2017}
}

@article{DSF,
  title={Diverse trajectory forecasting with determinantal point processes},
  author={Yuan, Ye and Kitani, Kris},
  journal={International Conference on Learning Representations},
  year={2020}
}

@inproceedings{BoM,
  title={Accurate and diverse sampling of sequences based on a “best of many” sample objective},
  author={Bhattacharyya, Apratim and Schiele, Bernt and Fritz, Mario},
  booktitle={Proceedings of the IEEE Conference on Computer Vision and Pattern Recognition},
  pages={8485--8493},
  year={2018}
}

@inproceedings{DLow,
  title={Dlow: Diversifying latent flows for diverse human motion prediction},
  author={Yuan, Ye and Kitani, Kris},
  booktitle={Computer Vision--ECCV 2020: 16th European Conference, Glasgow, UK, August 23--28, 2020, Proceedings, Part IX 16},
  pages={346--364},
  year={2020},
  organization={Springer}
}

@inproceedings{GSPS,
  title={Generating smooth pose sequences for diverse human motion prediction},
  author={Mao, Wei and Liu, Miaomiao and Salzmann, Mathieu},
  booktitle={Proceedings of the IEEE/CVF International Conference on Computer Vision},
  pages={13309--13318},
  year={2021}
}

@inproceedings{MOJO,
  title={We are more than our joints: Predicting how 3d bodies move},
  author={Zhang, Yan and Black, Michael J and Tang, Siyu},
  booktitle={Proceedings of the IEEE/CVF Conference on Computer Vision and Pattern Recognition},
  pages={3372--3382},
  year={2021}
}

@inproceedings{DivSamp,
  title={Diverse human motion prediction via gumbel-softmax sampling from an auxiliary space},
  author={Dang, Lingwei and Nie, Yongwei and Long, Chengjiang and Zhang, Qing and Li, Guiqing},
  booktitle={Proceedings of the 30th ACM international conference on multimedia},
  pages={5162--5171},
  year={2022}
}

@inproceedings{STARS,
  title={Diverse human motion prediction guided by multi-level spatial-temporal anchors},
  author={Xu, Sirui and Wang, Yu-Xiong and Gui, Liang-Yan},
  booktitle={European Conference on Computer Vision},
  pages={251--269},
  year={2022},
  organization={Springer}
}

@inproceedings{MotionDiff,
  title={Human joint kinematics diffusion-refinement for stochastic motion prediction},
  author={Wei, Dong and Sun, Huaijiang and Li, Bin and Lu, Jianfeng and Li, Weiqing and Sun, Xiaoning and Hu, Shengxiang},
  booktitle={Proceedings of the AAAI Conference on Artificial Intelligence},
  volume={37},
  number={5},
  pages={6110--6118},
  year={2023}
}

@inproceedings{Belfusion,
  title={Belfusion: Latent diffusion for behavior-driven human motion prediction},
  author={Barquero, German and Escalera, Sergio and Palmero, Cristina},
  booktitle={Proceedings of the IEEE/CVF International Conference on Computer Vision},
  pages={2317--2327},
  year={2023}
}

@inproceedings{HumanMAC,
  title={Humanmac: Masked motion completion for human motion prediction},
  author={Chen, Ling-Hao and Zhang, Jiawei and Li, Yewen and Pang, Yiren and Xia, Xiaobo and Liu, Tongliang},
  booktitle={Proceedings of the IEEE/CVF international conference on computer vision},
  pages={9544--9555},
  year={2023}
}

@inproceedings{CoMotion,
  title={CoMusion: Towards Consistent Stochastic Human Motion Prediction via Motion Diffusion},
  author={Sun, Jiarui and Chowdhary, Girish},
  booktitle={European Conference on Computer Vision},
  pages={18--36},
  year={2024},
  organization={Springer}
}

@article{TransFusion,
  title={TransFusion: A practical and effective transformer-based diffusion model for 3d human motion prediction},
  author={Tian, Sibo and Zheng, Minghui and Liang, Xiao},
  journal={IEEE Robotics and Automation Letters},
  year={2024},
  volume={9},
  number={7},
  pages={6232-6239},
  publisher={IEEE}
}

@misc{SwiftDiff,
      title={Bayesian-Optimized One-Step Diffusion Model with Knowledge Distillation for Real-Time 3D Human Motion Prediction}, 
      author={Sibo Tian and Minghui Zheng and Xiao Liang},
      year={2024},
      eprint={2409.12456},
      archivePrefix={arXiv},
      primaryClass={cs.CV},
      url={https://arxiv.org/abs/2409.12456}, 
}

@article{MotionWavelet,
  title  = {MotionWavelet: Human Motion Prediction via Wavelet Manifold Learning},
  author = {Yuming Feng and Zhiyang Dou and Ling-Hao Chen and Yuan Liu and 
            Tianyu Li and Jingbo Wang and Zeyu Cao and Wenping Wang and 
            Taku Komura and Lingjie Liu},
  journal={arXiv preprint arXiv:2411.16964},
  year   = {2024}
}

@inproceedings{SLD-HMP,
  title={Learning semantic latent directions for accurate and controllable human motion prediction},
  author={Xu, Guowei and Tao, Jiale and Li, Wen and Duan, Lixin},
  booktitle={European Conference on Computer Vision},
  pages={56--73},
  year={2024},
  organization={Springer}
}

@article{SkeletonDiff,
  title={Nonisotropic Gaussian Diffusion for Realistic 3D Human Motion Prediction},
  author={Curreli, Cecilia and Muhle, Dominik and Saroha, Abhishek and Ye, Zhenzhang and Marin, Riccardo and Cremers, Daniel},
  journal={Proceedings of the IEEE/CVF Conference on Computer Vision and Pattern Recognition},
  year={2025}
}

@InProceedings{MotionMap,
  title = {MotionMap: Representing Multimodality in Human Pose Forecasting},
  author = {Hosseininejad, Reyhaneh and Shukla, Megh and Saadatnejad, Saeed and Salzmann, Mathieu and Alahi, Alexandre},
  booktitle = {Proceedings of the IEEE/CVF Conference on Computer Vision and Pattern Recognition (CVPR)},
  year = {2025},
  publisher = {IEEE/CVF}
}

@article{krishnan2015deep,
  title={Deep kalman filters},
  author={Krishnan, Rahul G and Shalit, Uri and Sontag, David},
  journal={arXiv preprint arXiv:1511.05121},
  year={2015}
}

@article{tong2025diverse,
  title={Diverse Human Motion Prediction via Sampling on Grassmann Manifold},
  author={Tong, Hanqing and Ding, Wenwen and Li, Qing and Ding, Chongyang},
  journal={Digital Signal Processing},
  pages={105539},
  year={2025},
  publisher={Elsevier}
}

@article{kalman1960new,
  title={A New Approach to Linear Filtering and Prediction Problems},
  author={Kalman, R. E.},
  journal={Journal of Basic Engineering},
  volume={82},
  number={1},
  pages={35--45},
  year={1960},
  publisher={American Society of Mechanical Engineers}
}

@article{iglesias2013ensemble,
  title={Ensemble {K}alman methods for inverse problems},
  author={Iglesias, Marco A and Law, Kody J H and Stuart, Andrew M},
  journal={Inverse Problems},
  volume={29},
  number={4},
  pages={045001},
  year={2013},
  publisher={IOP Publishing}
}

@article{zheng2025ensemble,
  title={Ensemble Kalman Diffusion Guidance: A Derivative-free Method for Inverse Problems},
  author={Zheng, Hongkai and Chu, Wenda and Wang, Austin and Kovachki, Nikola B. and Baptista, Ricardo and Yue, Yisong},
  journal={Transactions on Machine Learning Research},
  year={2025},
  month={May},
  issn={2835-8856}
}

@article{ionescu2013human3,
  title={{Human3.6M}: Large Scale Datasets and Predictive Methods for {3D} Human Sensing in Natural Environments},
  author={Ionescu, Catalin and Papava, Dragos and Olaru, Vlad and Sminchisescu, Cristian},
  journal={IEEE Transactions on Pattern Analysis and Machine Intelligence},
  volume={36},
  number={7},
  pages={1325--1339},
  year={2013},
  publisher={IEEE}
}

@article{humaneva,
  title={Humaneva: Synchronized video and motion capture dataset and baseline algorithm for evaluation of articulated human motion},
  author={Sigal, Leonid and Balan, Alexandru O and Black, Michael J},
  journal={International journal of computer vision},
  volume={87},
  number={1},
  pages={4--27},
  year={2010},
  publisher={Springer}
}

@inproceedings{sul1998synthesis,
  title={Synthesis of human motion using kalman filter},
  author={Sul, Chan-hee and Jung, Sung-Kwon and Wohn, Kwang-Yun},
  booktitle={Proceedings. International Conference on Computer Vision (Cat. No. 98CH36271)},
  pages={109--114},
  year={1998},
  organization={IEEE}
}

@inproceedings{liu2010human,
  title={Human motion tracking based on unscented Kalman filter in sports domain},
  author={Liu, Guoli and Tang, Xiaoou},
  booktitle={2010 20th International Conference on Pattern Recognition},
  pages={2628--2631},
  year={2010},
  organization={IEEE}
}

@article{huang2024recurrent,
  title={A Recurrent Neural Network Enhanced Unscented Kalman Filter for Human Motion Prediction},
  author={Huang, Zhen and Van Gool, Luc},
  journal={arXiv preprint arXiv:2402.08389},
  year={2024}
}

@Article{LugrisRealtime2023,
  author    = {Lugr{\'\i}s, Urbano and P{\'e}rez-Soto, Manuel and Michaud, Florian and Cuadrado, Javier},
  title     = {Human motion capture, reconstruction, and musculoskeletal analysis in real time},
  journal   = {Multibody System Dynamics},
  year      = {2023},
  volume    = {60},
  number    = {1},
  pages     = {3--25},
  month     = {oct},
  doi       = {10.1007/s11044-023-09938-0},
  publisher = {Springer},
}

@ARTICLE{LannanTKFAE2022,
  author    = {Lannan, Nate and Zhou, Le and Fan, Guoliang},
  journal   = {IEEE Access},
  title     = {Human Motion Enhancement via Tobit Kalman Filter-Assisted Autoencoder},
  year      = {2022},
  volume    = {10},
  pages     = {29233-29251},
  doi       = {10.1109/ACCESS.2022.3157605},
  issn      = {2169-3536}
}

@article{ahmed1974discrete,
  title={Discrete cosine transform},
  author={Ahmed, Nasir and Natarajan, T\_ and Rao, Kamisetty R},
  journal={IEEE Transactions on Computers},
  volume={100},
  number={1},
  pages={90--93},
  year={1974},
  publisher={IEEE}
}

@inproceedings{wu2024frequency,
  title={Frequency guidance matters: Skeletal action recognition by frequency-aware mixed transformer},
  author={Wu, Wenhan and Zheng, Ce and Yang, Zihao and Chen, Chen and Das, Srijan and Lu, Aidong},
  booktitle={Proceedings of the 32nd ACM International Conference on Multimedia},
  pages={4660--4669},
  year={2024}
}

@article{zhou2022joint,
  title={Joint optimization of kinematics and anthropometrics for human motion denoising},
  author={Zhou, Le and Lannan, Nate and Fan, Guoliang},
  journal={IEEE sensors journal},
  volume={22},
  number={5},
  pages={4386--4399},
  year={2022},
  publisher={IEEE}
}

@inproceedings{liu2024recurrent,
  title={A recurrent neural network enhanced unscented kalman filter for human motion prediction},
  author={Liu, Wansong and Tian, Sibo and Hu, Boyi and Liang, Xiao and Zheng, Minghui},
  booktitle={International Symposium on Flexible Automation},
  volume={87882},
  pages={V001T07A012},
  year={2024},
  organization={American Society of Mechanical Engineers}
}

@article{yuan2019diverse,
  title={Diverse trajectory forecasting with determinantal point processes},
  author={Yuan, Ye and Kitani, Kris},
  journal={arXiv preprint arXiv:1907.04967},
  year={2019}
}


\begin{center}
    \Large \textbf{Supplementary Material}
\end{center}
\vspace{2em}

\appendix

\section*{Contents}
\vspace{0.5em}
\noindent
\begin{tabular*}{\textwidth}{@{}l@{\extracolsep{\fill}}r@{}}
\hyperref[suppl:Steady-State Mean Square Error]{\textbf{A. Steady-State Mean Square Error Derivation for Kalman Filter}} & \textbf{\pageref{suppl:Steady-State Mean Square Error}} \\
\addlinespace[0.6em]
\hyperref[suppl:adaptive_snr]{\textbf{B. Adaptive Kalman Parameterization via SNR Estimation}} & \textbf{\pageref{suppl:adaptive_snr}} \\
\addlinespace[0.6em]
\hyperref[suppl:justification]{\textbf{C. Justification for the DCT-Kalman Refinement Module}} & \textbf{\pageref{suppl:justification}} \\
\addlinespace[0.6em]
\hyperref[sec:datasets]{\textbf{D. Datasets and Evaluation Metrics}} & \textbf{\pageref{sec:datasets}} \\
\addlinespace[0.6em]
\hyperref[sec:implementation]{\textbf{E. Implementation Details}} & \textbf{\pageref{sec:implementation}} \\
\addlinespace[0.6em]
\hyperref[sec:comparison_wavelet]{\textbf{F. Comparison to Wavelet-based Method}} & \textbf{\pageref{sec:comparison_wavelet}} \\
\addlinespace[0.6em]
\hyperref[sec:additional visualization]{\textbf{G. Additional Visualization Results}} & \textbf{\pageref{sec:additional visualization}} \\
\addlinespace[0.6em]
\hyperref[sec:angle_constraint_details]{\textbf{H. Joint Angle Constraint Details}} & \textbf{\pageref{sec:angle_constraint_details}} \\
\addlinespace[0.6em]
\hyperref[sec:algorithm]{\textbf{I. Algorithm}} & \textbf{\pageref{sec:algorithm}} \\
\addlinespace[0.6em]
\hyperref[sec:discussion]{\textbf{J. Discussion and Future Work}} & \textbf{\pageref{sec:discussion}} \\
\end{tabular*}

\vspace{3em}

\section{Steady-State Mean Square Error Derivation for Kalman Filter (Applied to DCT High-Frequency Denoising)}
\label{suppl:Steady-State Mean Square Error}

In our method, after obtaining the predicted motion sequences $\widehat{\mathbf{Y}}$ from the network, we transform the output to the frequency domain via the Discrete Cosine Transform (DCT). For each predicted high-frequency DCT coefficient sequence (indexed by frequency $k \geq k_0$ where $k_0$ is chosen based on the signal-to-noise characteristics), we model the relationship between adjacent frequency components as a first-order Gaussian-Markov process along the frequency index:
\begin{align}
    x_k &= x_{k-1} + w_k, \qquad w_k \sim \mathcal{N}(0, Q) \\
    z_k &= x_k + v_k, \qquad v_k \sim \mathcal{N}(0, R)
\end{align}
Here, $x_k$ is the true value of the DCT coefficient at frequency index $k$ (for a fixed joint and coordinate channel), $z_k$ is the corresponding predicted coefficient from the network, $w_k$ is process noise with variance $Q$, and $v_k$ is observation noise with variance $R$. We assume that $w_k$ and $v_k$ are mutually independent, zero-mean, and independent across frequency indices. The parameters $Q$ and $R$ are estimated from the training data or set based on prior knowledge about the spectral smoothness of motion (affecting $Q$) and the network prediction accuracy (affecting $R$).

The Kalman filter is applied to each DCT high-frequency sequence as follows. Define the estimate at frequency index $k$ as $\hat{x}_{k|k}$, and let $P_{k|k}$ be its associated mean square error (posterior error covariance). The recursive Kalman filter equations are:
\begin{align}
    \text{Prediction:} \quad & \hat{x}_{k|k-1} = \hat{x}_{k-1|k-1} \\
    & P_{k|k-1} = P_{k-1|k-1} + Q \\
    \text{Update:} \quad & K_k = \frac{P_{k|k-1}}{P_{k|k-1} + R} \\
    & \hat{x}_{k|k} = \hat{x}_{k|k-1} + K_k (z_k - \hat{x}_{k|k-1}) \\
    & P_{k|k} = (1 - K_k) P_{k|k-1}
\end{align}
where $K_k$ is the Kalman gain at frequency index $k$.

As $k \to \infty$, $P_{k|k}$ converges to a steady-state value $P^*$. In steady-state, we have $P_{k|k} = P_{k-1|k-1} = P^*$, so
\begin{align}
    P^* &= (1 - K^*)(P^* + Q),
\end{align}
where the steady-state Kalman gain is
\begin{equation}
    K^* = \frac{P^* + Q}{P^* + Q + R}.
\end{equation}

Substituting $K^*$ into the steady-state equation:
\begin{align}
    &P^* = \left[1 - \frac{P^* + Q}{P^* + Q + R}\right](P^* + Q) \\
    &~~~~~ = \frac{R}{P^* + Q + R}(P^* + Q) \\
     \Rightarrow & P^*(P^* + Q + R) = R(P^* + Q) \\
    &{P^*}^2 + QP^* + RP^* = RP^* + RQ \\
    &{P^*}^2 + QP^* = RQ \\
    &{P^*}^2 + QP^* - RQ = 0
\end{align}

Solving the quadratic equation for $P^*$ using the quadratic formula, we obtain
\begin{equation}
    P^* = \frac{-Q + \sqrt{Q^2 + 4QR}}{2}. 
\end{equation}
Since $P^* > 0$, we choose the positive root. Thus, $P^*$ gives the steady-state mean square error of the filtered high-frequency DCT coefficient sequence after Kalman filtering.

To verify that $P^* < R$ for any $Q, R > 0$, we need to show:
\begin{align}
    &\frac{-Q + \sqrt{Q^2 + 4QR}}{2} < R \\
   &\Longleftrightarrow -Q + \sqrt{Q^2 + 4QR} < 2R \\
   &\Longleftrightarrow \sqrt{Q^2 + 4QR} < 2R + Q.
\end{align}
Squaring both sides (noting that $2R + Q > 0$):
\begin{align}
    &Q^2 + 4QR < (2R + Q)^2 = 4R^2 + 4QR + Q^2 \\
    &\Longleftrightarrow  0 < 4R^2,
\end{align}
which is always true for $R > 0$. Therefore, $P^* < R$ is proven, demonstrating that the Kalman filter always reduces the mean square error compared to the raw prediction (unfiltered observation noise variance $R$).

Specifically, the filtering is optimal in the MMSE sense for linear Gaussian models, thus our method leverages the DCT-Kalman approach for optimal denoising of high-frequency components in a mathematically principled way. This analysis rigorously demonstrates that our post-processing pipeline (network prediction $\to$ DCT $\to$ Kalman filtering) yields a lower expected error in the high-frequency range than direct use of the network prediction or traditional non-adaptive low-pass filtering.

\begin{figure}[h]
    \centering
    \includegraphics[width=0.85\columnwidth]{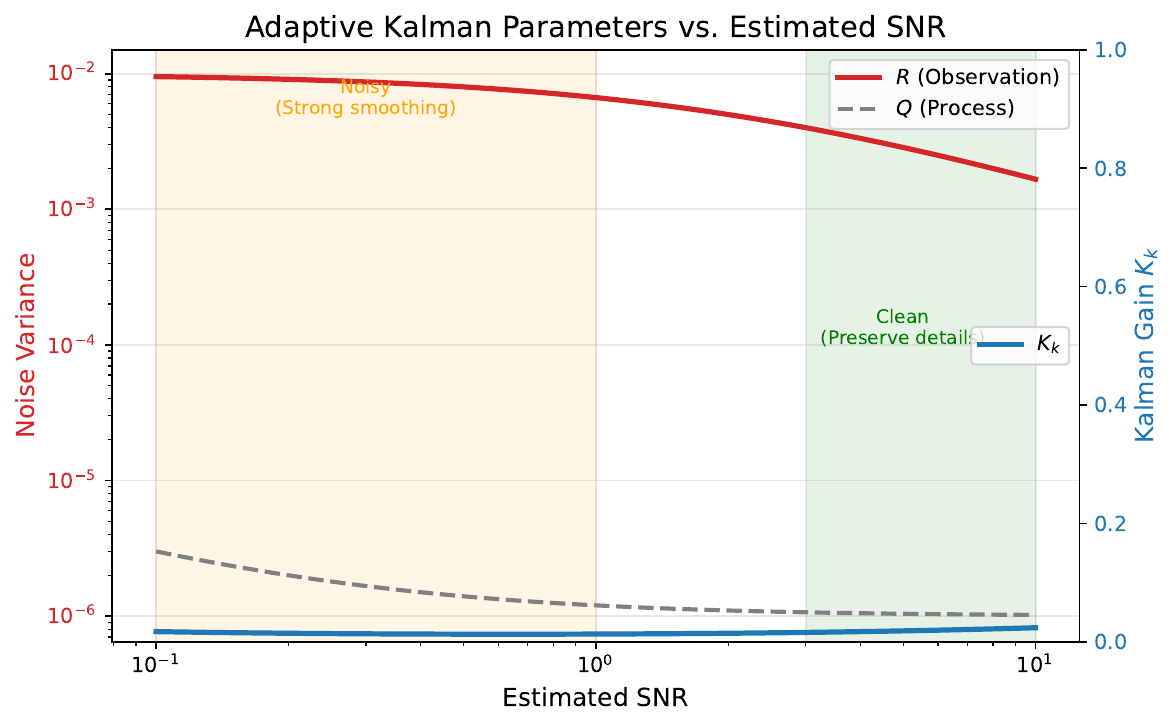}
    \caption{Behavior of adaptive Kalman filter parameters versus estimated SNR ($\text{SNR}_{\text{est}}$). As signal quality increases (higher SNR), observation noise $R$ decreases sharply, while process noise $Q$ decreases moderately. Consequently, the Kalman gain $K_k$ increases, shifting the filter's behavior from strong smoothing (low SNR, noisy signals) towards detail preservation (high SNR, clean signals).}
    \label{fig:adaptive_kalman_snr}
\end{figure}

To illustrate the behavior of our adaptive Kalman filtering strategy, we analyze how the key parameters respond to varying signal, represented by the estimated Signal-to-Noise Ratio ($\text{SNR}_{\text{est}}$). Figure~\ref{fig:adaptive_kalman_snr} plots adaptive noise variances ($Q$, $R$) and the resulting steady-state Kalman gain ($K_k$) against $\text{SNR}_{\text{est}}$. 
As shown, the observation noise $R$ decreases significantly as $\text{SNR}_{\text{est}}$ increases, reflecting higher confidence in cleaner signals (Eq.~\ref{eq:R}). The process noise $Q$ also decreases, enforcing stronger smoothness priors for higher quality signals (Eq.~\ref{eq:Q}). 
Importantly, the Kalman gain $K_k$ increases with higher $\text{SNR}_{\text{est}}$. This visualizes the adaptive mechanism: for noisy signals (low $\text{SNR}_{\text{est}}$), the low gain prioritizes the filter's internal prediction, resulting in stronger smoothing; for clean signals (high $\text{SNR}_{\text{est}}$), the high gain allows the filter to track the input coefficients, preserving fine-grained motion details.

\section{Adaptive Kalman Parameterization via SNR Estimation}
\label{suppl:adaptive_snr}

To enable adaptive smoothing, we estimate the Kalman filter noise parameters based on the spectral energy of the predicted motion. Let $\widehat{\mathbf{Y}} \in \mathbb{R}^{T' \times J \times 3}$ be a predicted sequence. We first compute its Discrete Cosine Transform (DCT) along the temporal axis for each joint $j \in \{1,\dots,J\}$ and coordinate channel $d \in \{1,2,3\}$, yielding frequency coefficients $c_k^{(j,d)}$.

The DCT conserves signal energy (Parseval’s Theorem), allowing us to analyze the energy distribution in the frequency domain. For each individual channel $(j,d)$, we define the high-frequency energy ratio $\rho^{(j,d)}$ as the ratio of energy beyond a cutoff frequency $k_0$ to the total signal energy:
\begin{equation}
\rho^{(j,d)} = \frac{\sum_{k \geq k_0} |c_k^{(j,d)}|^2}{\sum_{k=0}^{K-1} |c_k^{(j,d)}|^2 + \epsilon},
\end{equation}
where $\epsilon = 10^{-8}$ ensures numerical stability. This per-channel ratio $\rho^{(j,d)}$ serves as a precise, localized measure of jitter or instability.

From this ratio, we estimate the channel's Signal-to-Noise Ratio (SNR) by treating the low-frequency energy ($1-\rho$) as the "signal" and the high-frequency energy ($\rho$) as the "noise":
\begin{equation}
\text{SNR}_{\text{est}} = \frac{1 - \rho^{(j,d)}}{\rho^{(j,d)} + \epsilon}.
\end{equation}

The adaptive process noise $Q$ and observation noise $R$ are then set as non-linear functions of this estimated SNR, matching the formulation in Section 4.3 of the main paper:
\begin{align}
Q &= Q_0 \left(1 + \frac{\lambda_Q}{\text{SNR}_{\text{est}} + \epsilon}\right), \label{eq:Q} \\
R &= \frac{R_0}{1 + \lambda_R \cdot \text{SNR}_{\text{est}}}, \label{eq:R} 
\end{align}
where $Q_0, R_0$ are the base noise variances, and $\lambda_Q, \lambda_R$ are sensitivity hyperparameters.

\noindent\textbf{Kalman Gain Interpretation.}
This formulation has a clear physical interpretation.

First, for noisy signals (low $\text{SNR}_{\text{est}}$): The observation noise $R$ is set to a large value (approaching $R_0$), indicating low trust in the noisy observation. The process noise $Q$ also becomes large, allowing the filter to remain agile. The large $R$ dominates the Kalman gain calculation, resulting in a small gain $K_k$. This small gain causes the filter update step to heavily favor its internal smooth prediction over the noisy observation, thus applying strong smoothing. 

Second, for clean signals (high $\text{SNR}_{\text{est}}$): The observation noise $R$ becomes very small, reflecting high trust in the observation. The process noise $Q$ approaches its base $Q_0$, enforcing smoothness. The small $R$ results in a larger Kalman gain $K_k$, allowing the filter to closely track the (trusted) input signal while conservatively preserving its details.

\noindent\textbf{Steady-State Error Analysis.}
The steady-state mean square error of the estimate, $P^*$, for a channel with a given $\text{SNR}_{\text{est}}$ still satisfies the algebraic Riccati equation ~\cite{kalman1960new}:
\begin{equation}
{P^*}^2 + Q P^* - Q R = 0,
\end{equation}
where $Q \equiv Q(\text{SNR}_{\text{est}})$ and $R \equiv R(\text{SNR}_{\text{est}})$ are the adaptive parameters defined above. The valid solution for the error covariance is:
\begin{equation}
P^*(\text{SNR}_{\text{est}}) = \frac{-Q + \sqrt{Q^2 + 4QR}}{2}.
\end{equation}
This explicitly shows that the filter's steady-state estimation error $P^*$ is dynamically controlled by the spectral properties of the input signal via the estimated SNR.

\noindent\textbf{Filtering Control Function.}
The steady-state Kalman gain $K^*$ itself can be viewed as a data-driven control function $K^*(\text{SNR}_{\text{est}})$:
\begin{equation}
K^*(\text{SNR}_{\text{est}}) = \frac{P^*(\text{SNR}_{\text{est}}) + Q}{P^*(\text{SNR}_{\text{est}}) + Q + R}.
\end{equation}
This gain $K^*$ is small for low SNR signals and larger for high SNR signals, ensuring a smooth, adaptive transition from aggressive smoothing for noisy motions to gentle filtering for clean ones.

\section{Justification for the DCT-Kalman Refinement Module}
\label{suppl:justification}
Our choice of an adaptive Kalman filter in the Discrete Cosine Transform (DCT) domain is principled, aiming to achieve efficient and signal-aware suppression. We perform filtering in the frequency domain, where low-frequency coefficients encode global trajectory trends and high-frequency components capture rapid local oscillations—typically perceived as motion jitter. Among frequency transforms, the Discrete Cosine Transform (DCT) is particularly suitable for motion signals due to its strong energy compaction and real-valued orthogonality, which enable more distinct localization of jitter-related components than the DFT or wavelet bases~\cite{MotionWavelet}. Rather than simply discarding high-frequency coefficients, which can remove fine-grained dynamics, we model the sequence of DCT coefficients as a pseudo-temporal process and apply a Kalman filter. This formulation treats frequency indices as a sequential ordering, applying recursive estimation \textit{along the spectral axis} to suppress stochastic variance while preserving deterministic spectral structures.

It is important to distinguish this from the non-linear Kalman filters (e.g., UKF, EKF), which model complex \textit{temporal} dynamics~\cite{LugrisRealtime2023, huang2024recurrent, liu2024recurrent}. Our approach, in contrast, applies a simple \textit{linear} state-space model ($x_k = x_{k-1} + w_k$, $z_k = x_k + v_k$) directly to the \textit{frequency} coefficients. This linear model is justified for two primary reasons: (1) The underlying "clean" frequency spectrum of natural motion is assumed to be smooth, lacking abrupt discontinuities between neighboring components. Therefore, modeling the relationship between adjacent coefficients with a first-order Gaussian-Markov model ($x_k \approx x_{k-1}$) is a reasonable and well-founded assumption. (2) The critical non-linearity in our system is not in the state transition, but is instead handled by our adaptive parameterization strategy, which dynamically adjusts $Q$ and $R$ based on signal properties.

The key novelty lies in adaptive modulation. Unlike fixed-parameter Kalman filters, which risk over-smoothing clean motions or under-smoothing noisy ones, our adaptive variant dynamically scales the process and observation noise ($Q$, $R$) according to the estimated spectral energy ratio in the high-frequency band. Consequently, the adaptive DCT-Kalman module achieves selective artifact suppression—aggressively reducing stochastic jitter in degraded predictions while retaining the natural temporal fidelity of clean motions.

\section{Datasets and Evaluation Metrics}
\label{sec:datasets}

\subsection{Datasets}
We validate our framework on two widely recognized public benchmarks for human motion prediction.

\noindent\textbf{Human3.6M~\cite{ionescu2013human3}.} This large-scale benchmark contains 3.6 million 3D human motion frames, featuring 11 subjects across 15 actions, recorded at 50 Hz. We adhere to the standard protocol~\cite{SLD-HMP, DivSamp}, utilizing subjects S1, S5, S6, S7, and S8 for training and S9, S11 for testing. The evaluation task involves predicting 100 future frames from 25 observed frames using a 17-joint skeleton representation.

\noindent\textbf{HumanEva-I~\cite{humaneva}.} This dataset serves as another crucial, albeit smaller, benchmark for evaluation. It includes 3 subjects performing 5 actions, recorded at 60 Hz. We follow the official training and testing splits. The objective is to forecast 60 future frames given a history of 15 frames with a 15-joint skeleton representation.

\subsection{Evaluation Metrics}
Following prior works~\cite{SLD-HMP, MotionWavelet}, we employ a comprehensive set of metrics to assess both the accuracy and diversity of the $K$ generated predictions $\{\widehat{\mathbf{Y}}_k\}_{k=1}^K$.

\noindent\textbf{Average Pairwise Distance (APD).} To quantify the diversity of our predictions, we use APD, which calculates the average $L_1$ distance across all pairs of generated motion samples: $\frac{2}{K(K-1)}\sum_{j=1}^{K}\sum_{k=j+1}^{K}||\widehat{\mathbf{Y}}_j - \widehat{\mathbf{Y}}_k||_1$. A higher value indicates greater diversity.

\noindent\textbf{Average Displacement Error (ADE).} For overall trajectory accuracy, we report ADE. It is defined as the time-averaged $L_2$ distance between the ground truth (GT) motion $Y$ and the geometrically closest prediction from the generated set: $\min_{k} \frac{1}{T_f} \sum_{t=1}^{T_f} ||\mathbf{Y}_t - \widehat{\mathbf{Y}}_{k,t}||_2$.

\noindent\textbf{Final Displacement Error (FDE).} Complementing ADE, the FDE metric focuses specifically on the accuracy at the final timestep of the prediction horizon, computed as: $\min_{k} ||\mathbf{Y}_{T_f} - \widehat{\mathbf{Y}}_{k, T_f}||_2$.

\noindent\textbf{MMADE and MMFDE.} To evaluate performance in a multi-modal context, we report Multi-Modal ADE and FDE. These metrics compare the set of predictions against multiple plausible ground truth sequences $\{\mathbf{Y}_m\}_{m=1}^M$, which are collected from the dataset based on shared motion context with the input. Specifically, MMADE is computed as $\frac{1}{M}\sum_{m=1}^{M}\min_{k} \allowbreak \frac{1}{T_f}\sum_{t=1}^{T_f}||\mathbf{Y}_{m,t} - \widehat{\mathbf{Y}}_{k,t}||_2$, and MMFDE is computed as $\frac{1}{M}\sum_{m=1}^{M}\min_{k} \allowbreak ||\mathbf{Y}_{m,T_f} - \widehat{\mathbf{Y}}_{k,T_f}||_2$. The multi-modal ground truth is defined as $\{\mathbf{Y}_m\}_{m=1}^M = \{\mathbf{Y}_m \mid \|\mathbf{X}-\mathbf{X}_m\|_2 \leq \epsilon\}$, obtained by clustering future motions with similar past motion patterns within a distance threshold $\epsilon$~\cite{yuan2019diverse}, where $\mathbf{X}$ is the observed past motion, $\mathbf{Y}_m$ are the future motions, and $\mathbf{X}_m$ are the corresponding past motions of $\mathbf{Y}_m$.

\section{Implementation Details}
\label{sec:implementation}

\subsubsection{Network Architecture.} Our framework is built directly upon the architectural design of SLD-HMP~\cite{SLD-HMP}. The encoder (E) and decoder (D) are each composed of 4 Spatio-Temporal Graph Convolutional Network (STGCN) layers. The channel dimensions progress as (3, 128, 64, 128, 128) for E and (384, 128, 64, 128, 3) for D. The Query to Latent Projection (QLP) module consists of 3 STGCN layers. The semantic latent basis has a final dimension of $30 \times 256$, and we utilize the first 20 DCT coefficients for the motion representation.

\subsubsection{Training Hyperparameters.} The model is trained end-to-end for 500 epochs with a batch size of 16 using the Adam optimizer. We set the initial learning rate to $1 \times 10^{-3}$ and employ a cosine decay scheduler. Our Frequency-Kalman refinement module is a post-processing step applied only during inference, with its hyperparameters set to $k_0=10, Q_0=1e-6, R_0=1e-2, \lambda_Q=0.2, \lambda_R=0.5$.

\subsubsection{Loss Function Formulation}
\label{sec:loss_formulation}

Our total training objective, $\mathcal{L}_{\text{total}}$, is a comprehensive formulation that integrates a variational objective adapted from our baseline, SLD-HMP~\cite{SLD-HMP}, with our novel physical validity constraints. The final objective is defined as:
\begin{equation}
\mathcal{L}_{\text{total}} = \mathcal{L}_{\text{base}} + \lambda_{\text{temporal}} \mathcal{L}_{\text{temporal}} + \lambda_{\text{angle}} \mathcal{L}_{\text{angle}}.
\end{equation}

\noindent\textbf{Variational Objective.} The core of our training is the variational objective, $\mathcal{L}_{\text{base}}$, which is designed to ensure both accuracy and diversity. It combines several terms, whose formulations follow~\cite{SLD-HMP}. The full objective is defined as:
\begin{equation}
\begin{split}
    \mathcal{L}_{\text{base}} = {} & \lambda_{\text{recon}} \mathcal{L}_{\text{recon}} +  \lambda_{\text{h}} \mathcal{L}_{\text{h}} + \lambda_{\text{mm}} \mathcal{L}_{\text{mm}} \\
    & + \lambda_{\text{d}} \mathcal{L}_{d} + \lambda_{\text{limb}} \mathcal{L}_{\text{limb}} + \lambda_{\text{nf}} \mathcal{L}_{\text{nf}}.
\end{split}
\end{equation}
The components are as follows:
\begin{itemize}
    \item \textbf{Reconstruction Loss ($\mathcal{L}_{\text{recon}}$)}: This is the primary accuracy term, which minimizes the $L_2$ distance between the ground truth future motion $Y$ and the closest prediction $\widehat{\mathbf{Y}}_k$ among the $K$ generated samples, defined as $\min_{k} ||\widehat{\mathbf{Y}}_{k} - \mathbf{Y}||^2$.
    
    \item \textbf{Historical Reconstruction Loss ($\mathcal{L}_{\text{h}}$)}: Ensures temporal consistency by penalizing the error in reconstructing the observed past motion $X$. It is formulated as $\frac{1}{K}\sum_{k=1}^{K}||\widehat{\mathbf{X}}_{k} - \mathbf{X}||^2$.
    
    \item \textbf{Multimodal Reconstruction Loss ($\mathcal{L}_{\text{mm}}$)}: Promotes coverage of the data distribution by extending the reconstruction loss to a set of $M$ plausible ground truth futures $\{\mathbf{Y}_m\}$, defined as $\frac{1}{M}\sum_{m=1}^{M}\min_{k}||\widehat{\mathbf{Y}}_{k} - \mathbf{Y}_{m}||^2$.
    
    \item \textbf{Diversity-Promoting Loss ($\mathcal{L}_{d}$)}: Encourages the model to generate distinct futures by penalizing similarity between pairs of predictions, normalized across all pairs: $\frac{2}{K(K-1)}\sum_{j=1}^{K}\sum_{k=j+1}^{K}\exp(-\frac{||\widehat{\mathbf{Y}}_{j} - \widehat{\mathbf{Y}}_{k}||_{1}}{\alpha})$.
    
    \item \textbf{Limb Length Loss ($\mathcal{L}_{\text{limb}}$)}: A physical constraint that penalizes deviations between the limb lengths of the prediction $\hat{L}_k$ and the ground truth $L$, ensuring anatomical integrity: $\frac{1}{K}\sum_{k=1}^{K}||\hat{L}_{k} - L||^2$.
    
    \item \textbf{Pose Prior Loss ($\mathcal{L}_{\text{nf}}$)}: Measures the likelihood of the generated poses $\widehat{\mathbf{Y}}_k^{pose}$ using a normalizing flow $p_{nf}$ to ensure predicted poses are realistic: $-\sum_{k=1}^{K}\log p_{nf}(\widehat{\mathbf{Y}}_k^{pose})$.

    \item \textbf{Physical Constraint Losses ($\mathcal{L}_{\text{temporal}}$, $\mathcal{L}_{\text{angle}}$)}: These two terms encourage temporally smooth and anatomically plausible motion by penalizing abrupt frame-to-frame changes and joint angle violations. Their detailed formulations and joint-wise angle ranges are provided in Sec.4.2 from the main paper and Sec.~\ref{sec:angle_constraint_details} from the Supplementary Material.
\end{itemize}

\subsection{Loss Weights.} To balance the contributions of different objectives, the weights ($\lambda$) for each loss term are configured specifically for each dataset, as detailed in Table~\ref{tab:loss_weights}.

\begin{table}[h]
\centering
\small
\caption{Loss weights configuration.}
\label{tab:loss_weights}
\begin{tabular}{lcc}
\toprule 
\textbf{Loss Term} & \textbf{SLD-HMP} & \textbf{Ours} \\
\midrule
$\lambda_{\text{recon}}$ & 8.0 & 11.0 \\
$\lambda_{\text{h}}$ & 10.0 & 16.0 \\
$\lambda_{\text{mm}}$ & 4.0 & 0.1 \\
$\lambda_{\text{d}}$ & 16.0 & 0.63 \\
$\lambda_{\text{limb}}$ & 50.0 & 0.5 \\
$\lambda_{\text{nf}}$ & 0.002 & 0.002 \\
\bottomrule 
\end{tabular}
\end{table}
\vspace{-0.5in}
\section{Comparison to Wavelet-based Method}
\label{sec:comparison_wavelet}

To provide a comprehensive evaluation against other frequency-based methods, we display a direct comparison with MotionWavelet~\cite{MotionWavelet}, a recent work that leverages wavelet transformations for human motion 
prediction. The wavelet transform provides a multi-resolution analysis that captures both time and frequency information simultaneously, decomposing signals into approximation (low-frequency) and detail (high-frequency) coefficients at multiple scales. MotionWavelet applies 2D DWT (Discrete Wavelet Transform) along temporal and spatial dimensions to construct a motion manifold, then trains a diffusion model within this space and employs manifold-shaping guidance during sampling. In contrast, KHMP uses DCT for frequency decomposition and applies an adaptive Kalman filter with SNR-based parameter adjustment ($Q$ and $R$) to each prediction's noise characteristics. 

As shown in Table~\ref{tab:comparison_motionwavelet}, KHMP demonstrates superior performance across both datasets, outperforming MotionWavelet in most metrics. The key distinction lies in adaptivity: wavelet manifold guidance applies fixed iDWT-DWT transformations uniformly to maintain manifold structure, whereas our Kalman filter dynamically adjusts noise parameters ($Q$ and $R$) based on an estimated SNR for each prediction. For high-frequency jitter (low SNR), the filter increases $Q$ and $R$ to apply aggressive smoothing; for clean predictions (high SNR), it reduces these parameters to preserve motion details. This prediction-specific adaptation prevents over-smoothing of accurate predictions while suppressing noisy outputs, a mechanism that uniform transformations cannot achieve. 
\vspace{-0.2in}
\begin{table}[h]
\centering
\renewcommand\arraystretch{1.1}
\caption{Quantitative comparison between KHMP and MotionWavelet on HumanEva-I and Human3.6M datasets. Best results are highlighted in bold.}
\label{tab:comparison_motionwavelet}
\resizebox{\textwidth}{!}{
\begin{tabular}{l|ccccc|ccccc}
\toprule
\multirow{2}{*}{\textbf{Method}} & \multicolumn{5}{c|}{\textbf{HumanEva-I}} & \multicolumn{5}{c}{\textbf{Human3.6M}} \\
\cmidrule(lr){2-6} \cmidrule(lr){7-11}
& APD$\uparrow$ & ADE$\downarrow$ & FDE$\downarrow$ & MMADE$\downarrow$ & MMFDE$\downarrow$ & APD$\uparrow$ & ADE$\downarrow$ & FDE$\downarrow$ & MMADE$\downarrow$ & MMFDE$\downarrow$ \\
\midrule
MotionWavelet~\cite{MotionWavelet} & 4.171 & 0.235 & 0.213 & 0.304 & \textbf{0.280} & 6.506 & 0.376 & \textbf{0.408} & 0.466 & \textbf{0.443} \\
\textbf{KHMP (Ours)} & \textbf{7.481} & \textbf{0.188} & \textbf{0.204} & \textbf{0.301} & 0.291 & \textbf{9.235} & \textbf{0.349} & 0.441 & \textbf{0.436} & 0.468 \\
\bottomrule
\end{tabular}
}
\end{table}
\vspace{-0.3in}
\section{Additional Visualization Results}
\label{sec:additional visualization}

\begin{figure}[h]
    \centering
    \includegraphics[width=0.8\linewidth]{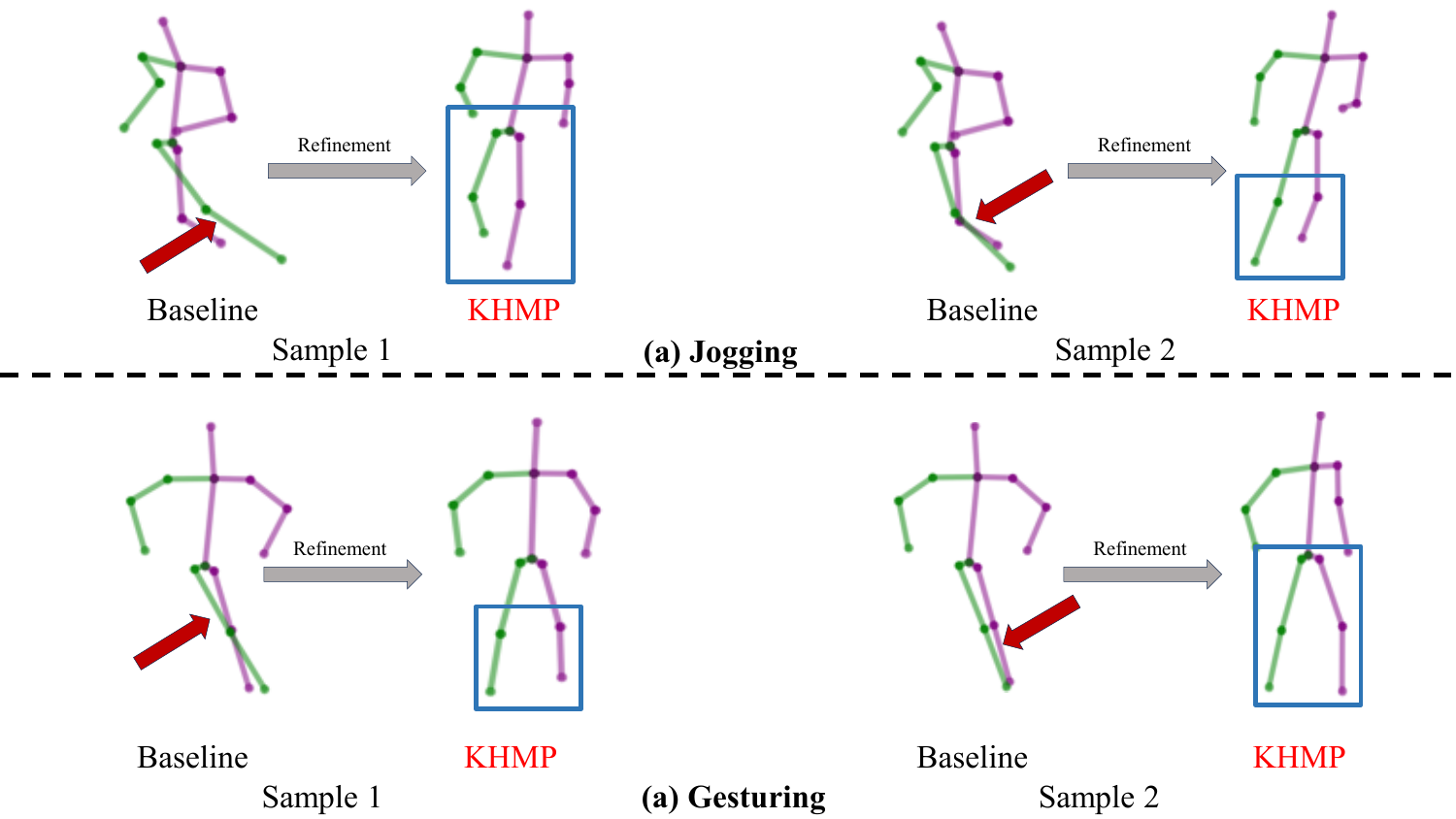} 
    \caption{Additional qualitative comparison between Baseline and KHMP predictions for selected frames from \textit{Jogging} and \textit{Gesturing} actions. Red arrows highlight implausible poses in the baseline, while blue boxes show the refined and more realistic results from KHMP.}
    \label{fig:suppl_qualitative_comparison}
\end{figure}

To further illustrate the qualitative improvements provided by our KHMP framework, Figure~\ref{fig:suppl_qualitative_comparison} presents additional visual comparisons against the baseline method. These examples focus on specific frames from predicted frames for the \textit{Jogging} and \textit{Gesturing} actions. As indicated by the red arrows, the baseline predictions exhibit unnatural limb configurations or jitter. In contrast, the KHMP results, highlighted by blue boxes, demonstrate enhanced physical plausibility with more natural skeletons.

\section{Joint Angle Constraint Details}
\label{sec:angle_constraint_details}

\subsection{Angle Computation}
For each joint $j$, we compute the joint angle $\theta_{t,j}$ at timestep $t$ using a triplet of connected joints or planes. Given the corresponding 3D positions or plane normals, we define vectors $\mathbf{v}_1$ and $\mathbf{v}_2$ representing the geometric relationship. The cosine of the joint angle is computed via the normalized dot product:
\begin{equation}
\cos(\theta_{t,j}) = \frac{\mathbf{v}_1 \cdot \mathbf{v}_2}{\|\mathbf{v}_1\| \|\mathbf{v}_2\| + \epsilon},
\end{equation}
where $\epsilon = 10^{-8}$ ensures numerical stability. This formulation is both differentiable and computationally efficient.

\subsection{Biomechanical Angle Constraints}
We define physiologically plausible angle ranges $[\theta_j^{\min}, \theta_j^{\max}]$ for each joint based on biomechanical studies and validated against training data to ensure they reflect natural motion patterns. We distinguish between two types of angular constraints:

\begin{itemize}
\item \textbf{Joint angles}: Angles between two connected bones (e.g., Head2Neck measures the angle between the head and neck segments).
\item \textbf{Plane angles}: Angles between a bone or body segment and an anatomical plane (e.g., Leg2HipPlane measures the deviation of the leg from the sagittal plane defined by the hip joints).
\end{itemize}

\subsection{Implementation}
The joint angle loss applies soft penalties only when angles violate the predefined limits. For each predicted pose $\widehat{\mathbf{Y}}_t$ at timestep $t$, we compute $\cos(\theta_{t,j})$ and evaluate:
\begin{equation}
\ell_{t,j} = 
\begin{cases}
(\cos(\theta_{t,j}) - \cos^{\max}_j)^2 & \text{if } \cos(\theta_{t,j}) > \cos^{\max}_j \\
(\cos^{\min}_j - \cos(\theta_{t,j}))^2 & \text{if } \cos(\theta_{t,j}) < \cos^{\min}_j \\
0 & \text{otherwise}
\end{cases}
\end{equation}
The final loss is averaged: $\mathcal{L}_{\text{angle}} = \frac{1}{T' \cdot N_{\text{angles}}} \sum_{t,j} \ell_{t,j}$, where $N_{\text{angles}}$ denotes the number of joints.

\section{Algorithm}
\label{sec:algorithm}
To systematically present our refinement approach, Algorithm~\ref{alg:kalman_refinement} summarizes the complete procedure of adaptive frequency-domain Kalman filtering. The algorithm takes the predicted motion sequence $\widehat{\mathbf{Y}}$ from the base model and applies three sequential stages: (1) frequency domain transformation via 1D DCT along the temporal axis (Line 1-2), (2) adaptive high-frequency refinement with SNR-based Kalman parameterization (Line 4-21), and (3) inverse transformation to reconstruct the refined motion (Line 23-24). This refinement operates independently on each joint-coordinate channel, preserving low-frequency components while selectively filtering high-frequency artifacts. The detailed formulations of the SNR estimation, adaptive parameterization, and Kalman update rules are provided in the main paper.

\section{Discussion and Future Work}
\label{sec:discussion}
\noindent\textbf{Discussion.}
Our work introduces KHMP, a framework that enhances stochastic motion prediction by integrating physics-based constraints during training with adaptive frequency-domain Kalman refinement post-inference. A key advantage is improving the temporal coherence and physical plausibility of predictions without retraining. The adaptive Kalman filter, driven by SNR estimation, effectively suppresses jitter while preserving clean motion dynamics. We note two aspects for potential improvements: 

(1) Diversity Maintenance. While the refinement effectively enhances prediction fidelity, it primarily maintains rather than expands prediction diversity (APD). Our results show consistent APD improvements (from 6.516 to 7.481 on HumanEva-I, and from 8.217 to 9.235 on Human3.6M), indicating that the adaptive filtering preserves sample-specific characteristics rather than homogenizing predictions. This occurs because our frequency-selective approach operates only on high-frequency jitter components while leaving low-frequency motion dynamics (which encode diversity) intact. However, the diversity gains are modest compared to the accuracy improvements, suggesting opportunities for future work to optimize quality and diversity jointly.

(2) Computational Efficiency. Kalman filtering introduces computational overhead during inference, creating a trade-off between quality and speed. On HumanEva-I, the refinement increases inference time by approximately 1.2×, while on Human3.6M, the overhead is approximately 1.6× due to the larger number of joints. While acceptable for offline applications, this suggests opportunities for optimization, such as selective refinement or parallelization, to enable more efficient real-time deployment.


\noindent\textbf{Future Work.}
Promising directions include extending KHMP to long-term prediction scenarios (e.g., beyond 1000ms), where error accumulation is more critical, and optimizing the computational efficiency of the frequency-Kalman refinement for potential real-time applications. Finally, while our main objective in this work was to demonstrate that our dual-pronged strategy (physical constraints and frequency refinement) significantly enhances VAE-based generative models, validating the generality of our refinement module by applying it to other backbones, such as diffusion-based models, is a valuable future direction.

\begin{algorithm}
\caption{Adaptive Frequency-Domain Kalman Refinement}
\label{alg:kalman_refinement}
\textbf{Input:} Predicted motion $\widehat{\mathbf{Y}} \in \mathbb{R}^{T' \times J \times 3}$, 
frequency threshold $k_0$, base variances $Q_0, R_0$, sensitivity $\lambda_Q, \lambda_R$ \\
\textbf{Output:} Refined motion $\widehat{\mathbf{Y}}'$ 
\begin{algorithmic}[1]
\State \colorbox{gray!20}{/* --- 1. Frequency Domain Transformation --- */}
\State Apply 1D DCT along temporal axis: $c^{(j,d)} \gets \text{DCT}(\widehat{\mathbf{Y}}_{:,j,d})$ 
for all $j \in \{1,\dots,J\}, d \in \{x,y,z\}$
\State
\State \colorbox{gray!20}{/* --- 2. Adaptive High-Frequency Refinement --- */}
\State Initialize $c_{\text{refined}} \gets c$
\For{$j \gets 1$ \textbf{to} $J$}
    \For{$d \in \{x, y, z\}$}
        \State Set $c_{\text{refined}, k}^{(j,d)} \gets c_k^{(j,d)}$ 
        for $k = 0, \ldots, k_0-1$ \hfill /* Preserve low frequencies */
        \State
        \State $\rho^{(j,d)} \gets \displaystyle\frac{\sum_{k \geq k_0} |c_k^{(j,d)}|^2}{\sum_{k=0}^{T'-1} |c_k^{(j,d)}|^2 + \epsilon}$, \quad 
               $\text{SNR}_{\text{est}} \gets \displaystyle\frac{1 - \rho^{(j,d)}}{\rho^{(j,d)} + \epsilon}$ 
               \hfill /* Estimate SNR */
        \State
        \State $Q \gets Q_0 \left(1 + \dfrac{\lambda_Q}{\text{SNR}_{\text{est}} + \epsilon}\right)$, \quad 
               $R \gets \dfrac{R_0}{1 + \lambda_R \cdot \text{SNR}_{\text{est}}}$ 
               \hfill /* Adaptive parameters */
        \State
        \State $\hat{x}_{k_0|k_0} \gets c_{k_0}^{(j,d)}$, \quad 
               $P_{k_0|k_0} \gets R$ \hfill /* Initialize Kalman state */
        \For{$k \gets k_0+1$ \textbf{to} $T'-1$}
            \State $\hat{x}_{k|k-1} \gets \hat{x}_{k-1|k-1}$, \quad 
                   $P_{k|k-1} \gets P_{k-1|k-1} + Q$ \hfill /* Predict */
            \State $K_k \gets \dfrac{P_{k|k-1}}{P_{k|k-1} + R}$ \hfill /* Kalman gain */
            \State $\hat{x}_{k|k} \gets \hat{x}_{k|k-1} + K_k (c_k^{(j,d)} - \hat{x}_{k|k-1})$ 
                   \hfill /* Update */
            \State $P_{k|k} \gets (1 - K_k) P_{k|k-1}$
            \State $c_{\text{refined}, k}^{(j,d)} \gets \hat{x}_{k|k}$
        \EndFor
    \EndFor
\EndFor
\State
\State \colorbox{gray!20}{/* --- 3. Inverse Transformation --- */}
\State Apply 1D IDCT: $\widehat{\mathbf{Y}}'_{:,j,d} \gets \text{IDCT}(c_{\text{refined}}^{(j,d)})$ 
for all $j, d$ \hfill /* Reconstruct motion */
\State
\State \textbf{return} $\widehat{\mathbf{Y}}'$
\end{algorithmic}
\end{algorithm}


%
%
\end{document}